\def\paperlanguage{} %% English
\pdfoutput=1 % for arxiv

\documentclass[letterpaper, 10 pt, conference]{ieeeconf}  % Comment this line out if you need a4paper

\usepackage{bm}
\usepackage{cite}
\usepackage{flushend}
\include{preamble}

\newcommand{\argmin}{\mathop{\rm arg~min}\limits}

\IEEEoverridecommandlockouts                              % This command is only needed if
\title{\LARGE \textbf
  {
    \switchlanguage%
    {%
      RAMIEL: A Parallel-Wire Driven Monopedal Robot \\for High and Continuous Jumping
    }%
    {%
      パラレルワイヤ型一本脚ロボットRAMIELによる跳躍動作の実現
    }%
  }
}

\author{Temma Suzuki$^{1}$, Yasunori Toshimitsu$^{1}$, Yuya Nagamatsu$^{1}$, Kento Kawaharazuka$^{1}$,\\Akihiro Miki$^{1}$, Yoshimoto Ribayashi$^{1}$, Masahiro Bando$^{1}$, Kunio Kojima$^{1}$, \\Yohei Kakiuchi$^{1}$, Kei Okada$^{1}$, and Masayuki Inaba$^{1}$% <-this % stops a space
  \thanks{$^{1}$ The authors are with the Department of Mechano-Informatics, Graduate School of Information Science and Technology, The University of Tokyo, 7-3-1 Hongo, Bunkyo-ku, Tokyo, 113-8656, Japan.
    {\texttt\small [suzuki, toshimitsu, nagamatsu, kawaharazuka, miki, ribayashi, bando, kojima, kakiuchi, okada, inaba]@jsk.t.u-tokyo.ac.jp}
  }
}
\begin{document}

\maketitle
\thispagestyle{empty}
\pagestyle{empty}

%%%%%%%%%%%%%%%%%%%%%%%%%%%%%%%%%%%%%%%%%%%%%%%%%%%%%%%%%%%%%%%%%%%%%%%%%%%%%%%%
\begin{abstract}
  \switchlanguage%
  {%
    Legged robots with high locomotive performance have been extensively studied, and various leg structures have been proposed. 
  Especially, a leg structure that can achieve both continuous and high jumps is advantageous for moving around in a three-dimensional environment. 
  In this study, we propose a parallel wire-driven leg structure, which has one DoF of linear motion and two DoFs of rotation and is controlled by six wires, as a structure that can achieve both continuous jumping and high jumping. 
  The proposed structure can simultaneously achieve high controllability on each DoF, long acceleration distance and high power required for jumping.
  In order to verify the jumping performance of the parallel wire-driven leg structure, we have developed a parallel wire-driven monopedal robot, RAMIEL. 
  RAMIEL is equipped with quasi-direct drive, high power wire winding mechanisms and a lightweight leg, and can achieve a maximum jumping height of 1.6 m and a maximum of seven continuous jumps.
  }%
  {%
    走破性能が高い脚型ロボットはこれまで盛んに研究が行われており, 様々な脚構造が提案されてきた. 
    特に連続跳躍と高い跳躍を両立できる脚構造は立体的な環境内で動き回る上で有利である. 
    本研究では連続跳躍と高い跳躍を両立する脚構造として, 直動1自由度と回転2自由度を有する脚を6本のワイヤによって制御するパラレルワイヤ脚構造を提案する. 
    パラレルワイヤ脚構造は跳躍動作に求められる高い自由度、長い加速距離、高い出力を両立することができる。
    パラレルワイヤ脚構造の跳躍性能を検証するため, パラレルワイヤ型一本脚ロボットRAMIELを開発した.. 
    RAMIELは準ダイレクトドライブで大出力なワイヤ巻取り機構と, 軽量な脚が搭載されており, 最大重心跳躍高さ1.6mの跳躍と, 最大7回の連続跳躍が達成された可能である.
  
  }%
\end{abstract}

\section{Introduction}\label{sec:introduction}
\switchlanguage%
{%
Legged robots have a higher ability to overcome discrete footholds and steps than wheeled robots, and are expected to be applied to transportation of goods and patrolling. % Here is an example.
In particular, legged robots that can jump are being actively researched because they can move over more three-dimensional terrain\cite{haldane2017repetitive,kau2019stanford, boston2022atlas, arm2019spacebok, Zaitsev2015taub, matthew2014multimobat}. %Insert example here.

Several leg robots, \cite{Zaitsev2015taub, matthew2014multimobat}, have been developed so far that can perform a single jump with a CoG (center of gravity) jump height of \SI{3}{\metre} (the CoG jump height is the difference between the height of the CoG at the moment the foot leaves the ground and at the highest point during the jump). 
On the other hand, robots that specialize in single high jump are not equipped with actuators necessary for posture control and adjustment of the next leg position in order to reduce the weight of the machine. 
As a result, it is difficult for them to move with continuous jumps.
In addition, it is predicted that the weight of the robot will increase significantly and the jumping height will decrease if the actuator for posture control is installed in the legged robot specialized for single high jumping. 

\begin{figure}[t]
  \centering
  \includegraphics[width=1.0\columnwidth]{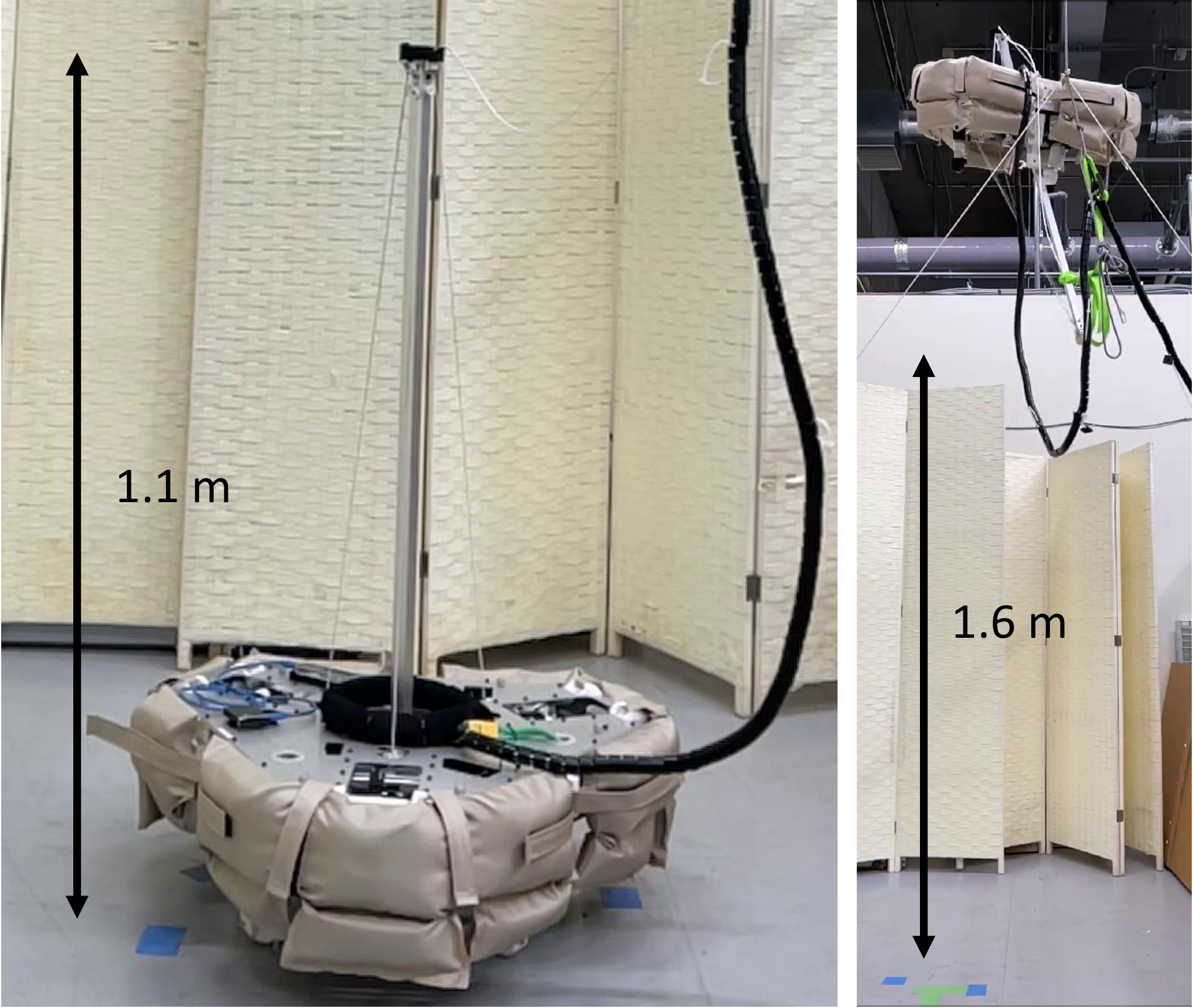}
  \vspace{-3.0ex}
  \caption{Overview of the parallel wire-driven leg and snapshot of \SI{1.6}{\metre} high jumping motion.}
  \label{figure:parallelwireoverview}
  \vspace{-1.0ex}
\end{figure}

Legged robots that can continuously jump and locomote have been developed, such as ATLAS \cite{boston2022atlas}, which achieves dynamic motions such as parkour using hydraulic drive, and Salto-1P \cite{haldane2017repetitive}, which achieves a jump of \SI{1}{\metre} by combining an electric motor, series elastic elements, and an optimized link structure. 
On the other hand, these continuous jumping robots have a maximum CoG jumping height of \SI{1.1}{\metre}, which is lower than that of legged robots specialized for single jumping \cite{haldane2017repetitive, kau2019stanford}. (For the previous study \cite{haldane2017repetitive, kau2019stanford}, which used the difference between the height of the CoG in a bent position and the height of the CoG at the highest point during a jump as the index of the jump height, the CoG was calculated by subtracting the leg stroke from the jump height described in the papers).

Therefore, the purpose of this study is to clarify the leg structure that can achieve both continuous jumping and high jumping.
As a leg structure to achieve this purpose, we propose a parallel wire-driven leg structure as shown in \figref{figure:parallelwireoverview}, in which a leg with one DoF of linear motion and two DoFs of rotation is controlled by six wires. 

In \secref{sec:parallel-wire-driven-leg}, we explain the parallel wire-driven leg structure and compare its jumping performance with that of existing leg structures. 
In \secref{sec:robot}, we describe the overall design of RAMIEL, a monopedal robot developed to verify the jumping performance of the parallel wire-driven leg structure and the detailed design of the wire winding mechanism.
In \secref{sec:controller}, we describe the hopping controller of RAMIEL.  
In \secref{sec:experiments}, we perform continuous jumping and high jumping experiments using RAMIEL and discuss the results of the experiments. 
In \secref{sec:conclusion}, we discuss the jumping performance of the parallel wire-driven leg structure based on the experimental results.
}%
{%
脚を用いて移動するロボットは台車型ロボットに比べ離散的な足場や段差などを乗り越える能力が高く, 物資運搬や巡回動作などへの応用が期待されている. %ここに例を入れる
特に跳躍して移動可能な脚ロボットは, より三次元的な地形に対応可能で走破性能が高く盛んに研究が行われている\cite{haldane2017repetitive,kau2019stanford, boston2022atlas, arm2019spacebok, Zaitsev2015taub, matthew2014multimobat}.%ここに例を入れる。

これまでに単発の重心跳躍高さ\SI{3}{\metre}程度の跳躍を行う脚ロボット\cite{Zaitsev2015taub, matthew2014multimobat}が複数開発されている(ここで言う重心跳躍高さとは, 足先が地面から離れた瞬間の重心の高さと, 跳躍中の最高点での重心の高さの差を表す). 
%BostonDynamics社のATLAS \cite{boston2022atlas,boston2022atlasjump}は油圧駆動の等身大ヒューマノイドで, \SI{0.5}{\metre}程度の箱への飛び乗り, 台の上からのバク宙, 溝の飛び越えなどが可能である.
%また, Haldaneらの開発したSalto-1P \cite{haldane2017repetitive}は\SI{0.1}{\kilogram}程度の小型ロボットで, 連続して\SI{1}{\metre}程度の跳躍を行い椅子や机に飛び乗ることができる.  
一方で\SI{3}{\metre}程度の単発の高い跳躍に特化したロボットは, 機体重量を軽くするために, 姿勢制御や次に脚をつく位置の調節に必要なアクチュエータを搭載していない. 
このため連続して跳躍を行い移動する能力に難がある.
そして高い跳躍に特化した脚ロボットに姿勢制御用のアクチュエータを搭載すると機体重量が大幅に増加し跳躍高さが低下することが予測される. 
%さらに単発の跳躍に特化した脚ロボットは脚部にバネやラッチなどの跳躍に必要な機構が集中し, 本体にはほとんど機構が搭載されないため全体重量に占める脚重量の割合が連続跳躍が可能な脚ロボットと比較して大きい. 
%全体重量に占める脚重量の割合が大きければ大きいほど脚を動かした際に本体が受ける反力が大きくなり, ロボットの姿勢が不安定になりやすい. 

連続して跳躍を行い移動することが可能な脚ロボットは, 油圧駆動を用いてパルクールなどのダイナミックな動作を実現したATLAS \cite{boston2022atlas}や, 電気モータと直列弾性要素と最適化されたリンク構造を組み合わせ\SI{1}{\metre}程度の跳躍を実現したSalto-1P \cite{haldane2017repetitive}などが開発されている. 
一方でこれらの連続跳躍が可能なロボットは重心跳躍高さが最大でも\SI{1.1}{\metre}程度となっており\cite{haldane2017repetitive, kau2019stanford}単発の跳躍に特化した脚ロボットに比べて重心跳躍高さが低くなっている(屈んだ状態での重心高さと跳躍中の最高点での重心の高さの差を跳躍高さの指標に用いている先行研究\cite{haldane2017repetitive, kau2019stanford}については, 論文に記載されていた跳躍高さから脚のストロークを差し引いて重心跳躍高さを計算した). 

そこで本研究では連続跳躍と高い跳躍を実現する脚構造を明らかにし, 脚ロボットの移動性能の向上に寄与することを目的とする.
そして本目的を達成する脚構造として\figref{figure:parallelwireoverview}に示すような, 直動1自由度と回転2自由度を有する脚を6本のワイヤを用いて制御するパラレルワイヤ脚構造を提案する. 
%そして連続跳躍と高い跳躍を両立する脚構造として, \figref{figure:parallelwireoverview}に示すようなパラレルワイヤ脚を提案する. 
\secref{sec:parallel-wire-driven-leg}ではパラレルワイヤ脚構造の説明と, 既存の脚構造との跳躍性能の比較を行う. 
\secref{sec:robot}ではパラレルワイヤ脚構造の跳躍性能を検証するために開発した一本脚ロボットRAMIELの全体設計とワイヤ巻取り機構の詳細設計について述べる.
\secref{sec:controller}ではRAMIELのホッピング制御器について説明する.  
\secref{sec:experiments}ではRAMIELを用いて連続跳躍実験と高い跳躍実験を行い, その実験結果について議論する. 
\secref{sec:conclusion}では実験結果を踏まえてパラレルワイヤ脚構造の跳躍性能について結論を述べる.
}%

\section{Structure and Advantages of Parallel Wire-driven Leg} \label{sec:parallel-wire-driven-leg}
\subsection{Structure of Parallel Wire-driven Leg}
\begin{figure}[t]
  \centering
  \includegraphics[width=1.0\columnwidth]{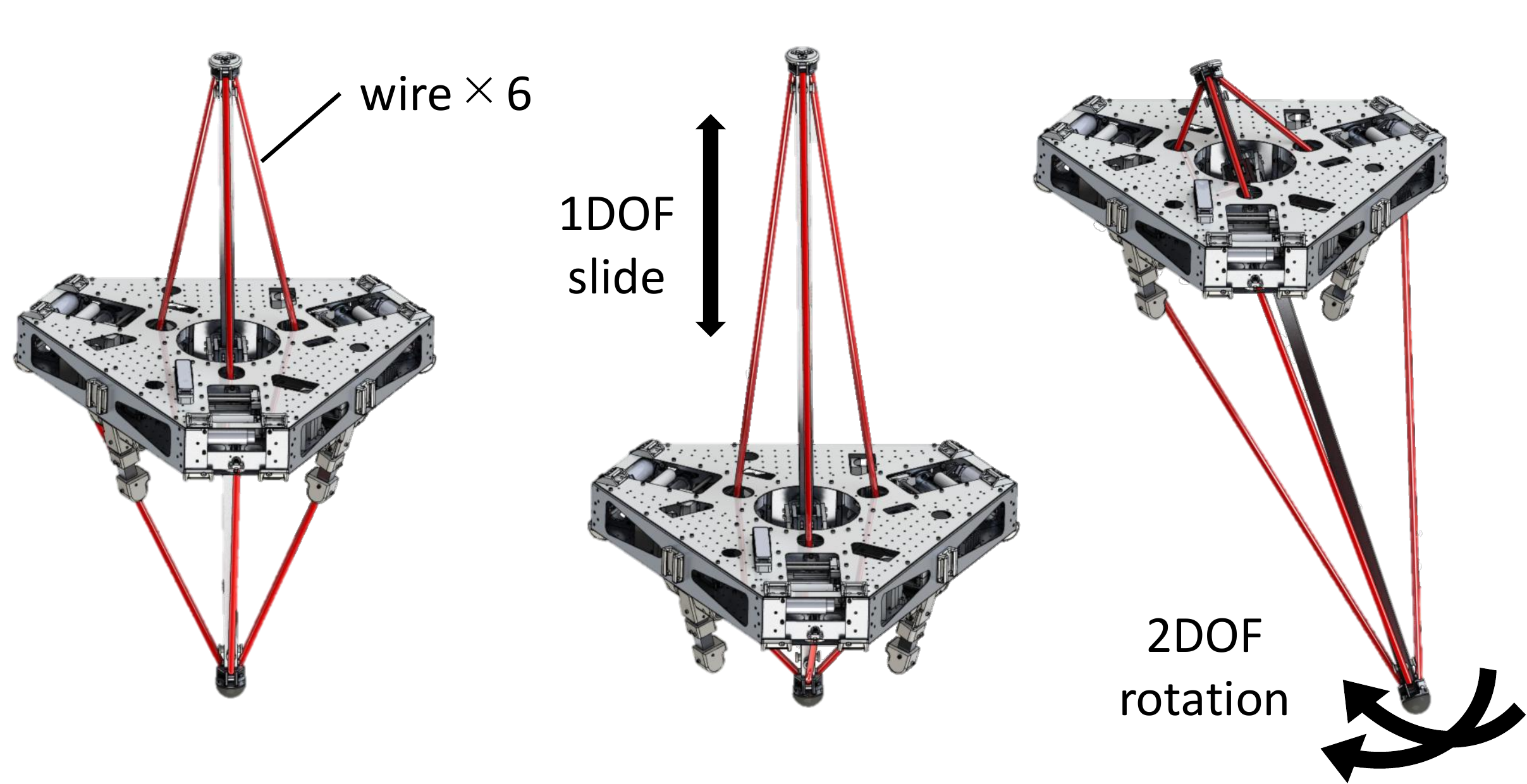}
  \vspace{-3.0ex}
  \caption{Structure of the parallel wire-driven leg.}
  \label{figure:parallelwirestructure}
  \vspace{-1.0ex}
\end{figure}

\switchlanguage%
{%
The parallel wire-driven leg structure is shown in \figref{figure:parallelwirestructure}. 
The parallel wire-driven leg structure consists of a body part and a leg part which has one DoF of linear motion and two DoFs of rotation relative to the body part. 
The joints of the leg are controlled by six wires extending from the main body.
Each wire is wound by an independent electric motor, and the target leg posture is realized by adjusting the winding length of the wire. 

A robot with a similar structure is FALCON\cite{kawamura1997falcon}.
However, FALCON is a robot designed for manipulation tasks whose body is fixed to the environment, and its design requirements are quite different from those of the legged robots that realize continuous jumping and high jumps. 
}%
{%
  パラレルワイヤ脚構造を\figref{figure:parallelwirestructure}に示す. 
  パラレルワイヤ脚構造は本体部分と, 本体部分に対して直動1自由度と回転2自由度を有する脚部分からなる. 
  脚の関節は本体から伸びる6本のワイヤを介して制御される.
  ワイヤはそれぞれ独立した電気モータによって巻き取られ, ワイヤを巻き取る長さを調節することで目標となる脚の姿勢を実現する. 
  
  同様の構造のロボットとしてFALCON \cite{kawamura1997falcon}が存在する.
  しかしFALCONは本体が環境に固定されたマニュピレーション用ロボットであり, 本研究で対象とする連続跳躍と高い跳躍を実現する脚ロボットとはその設計要件が大きく異なる. 
}%

\subsection{Advantages of Parallel Wire-driven Leg in Jumping}
\switchlanguage%
{%
  %In this chapter, we describe the advantages of the parallel wire-driven leg structure for high and continuous jumps.   
  In order for a legged robot to perform high jumps, the following three conditions must be met.
  \begin{enumerate}
    \item The force to lift the robot must be high.
    \item The acceleration distance must be large.
    \item The vertical velocity must not saturate during acceleration.
  \end{enumerate}
  Since the legged robot cannot accelerate vertically in the air, the jump height is uniquely determined from the vertical velocity at takeoff.
  In other words, to increase the jump height, we need to increase the velocity at take-off.
  In order to increase the velocity at takeoff, the acceleration of the robot must be large (the force to lift the robot must be large), and the acceleration distance must be large.
  In addition, the saturation of the output speed of the actuator must be considered.
  In an electric motor, as the angular velocity of the output axis increases, the back electromotive force also increases, so the angular velocity saturates at a certain value and further acceleration is not possible.
  Since actuators generally have such a maximum speed, it is also required that the actuator speed does not saturate during acceleration (vertical velocity must not saturate during acceleration) in order to increase the velocity at takeoff.

  In the animal-like rotational joint leg structure \cite{boston2022atlas, bledt2018cheetah3}, which is adopted by many legged robots, there is a trade-off between condition 2 (extension of acceleration distance) and condition 3 (prevention of velocity saturation).
  In the case of the leg that uses the rotational joint for vertical acceleration, it is necessary to take off with the knee extended in order to satisfy the condition 2.
  However, in the rotational joint leg structure, the more the knee is extended, the closer the leg is to the singularity, and the vertical velocity approaches zero.
  Therefore, in order to satisfy the condition 3, it is necessary to take off with the knee bent to some extent.
  This problem also occurs in legs with a special parallel link structure of non-animal type, such as the GOAT \cite{Kalouche2017GOAT}, if they are rotational joint legs.
  Thus, in the rotational joint leg structure, conditions 2 and 3 are contradictory, and the high jump is hindered.

  On the other hand, the parallel wire-driven leg structure does not have such conflict in conditions 2 and 3 as the rotating joint leg structure, and can satisfy the three conditions more easily.
  In the parallel wire-driven leg structure, the vertical speed does not decrease even when the leg is extended because the linear joint is used for vertical acceleration.
  Therefore, it is possible to accelerate with the maximum range of motion.
  In addition, the parallel wire-driven leg structure can lift the body by using three motors in parallel, which can easily satisfy the condition 1 compared with the leg structure using actuators in series \cite{boston2022atlas, bledt2018cheetah3}.

  Finally, we confirm the conditions required for the leg structure to perform continuous jumping.
  For continuous jumping, it is important that the leg is light.
  In continuous jumping, the robot moves its leg quickly to stabilize its posture.
  If the leg is heavy, the maximum acceleration of the leg is reduced and the posture of the main body easily collapses due to the recoil of the leg.
  The parallel wire-driven leg structure realizes a lightweight leg without any actuators and enables the robot to move the leg agilely. 

  The advantages of the parallel wire-driven leg structure in high jumps and continuous jumps have been described, but one of the disadvantages of the parallel wire-driven leg structure is that its self weight must be supported only by the actuator force.
  In the case of a leg with rotational joints, such as in animals, it is possible to support most of the self weight by extending the knee joint.
  On the other hand, parallel wire-driven legs have a disadvantage that they cannot support their own weight with their joints because they use vertical linear joints, and thus consume a lot of power just to keep standing.
  This disadvantage is the reverse of the advantage that the vertical velocity does not decrease even when the knees are extended, and is considered to be the result of specializing in jumping at the expense of power consumption.
}%
{%
  ここで高い跳躍と連続跳躍におけるパラレルワイヤ脚構造の利点を述べる.   
  
  脚ロボットが高い跳躍を行うためには以下の3つの条件を持たす必要がある.
  \begin{enumerate}
    \item ロボットを持ち上げる力が大きいこと
    \item 加速距離が大きいこと
    \item 垂直方向の速度が加速中に飽和しないこと
  \end{enumerate}
  脚ロボットは空中で垂直方向に加速することができないため, 離陸時の垂直方向の速度から跳躍高さが一意に定まる.
  つまり跳躍高さを上げるには離陸時の速度を大きくする必要がある.
  離陸時の速度を大きくするためには, ロボットの加速度を大きくすること(ロボットを持ち上げる力が大きいこと), 加速距離が大きいことが必要である.
  さらにアクチュエータの出力速度の飽和も考慮する必要がある.
  電気モータでは出力軸の角速度が増加すると逆起電力も増加するため, 角速度が一定値で飽和しそれ以上の加速ができない.
  アクチュエータには一般にこのような最大速度が存在するため, 離陸時の速度を上げるにはアクチュエータの速度が加速中に飽和しないこと(垂直方向の速度が加速中に飽和しないこと)も求められる.

  多くの脚ロボットが採用している動物のような回転関節型の脚構造\cite{boston2022atlas, bledt2018cheetah3}では, 条件2(加速距離の延長)と条件3(速度飽和の防止)がトレードオフの関係にある.
  回転関節を垂直方向の加速に用いる脚では, 条件2(加速距離の延長)を満たすために膝を伸ばした状態で離陸する必要がある.
  しかし回転関節脚構造では, 膝を伸ばすほど特異点に近づき, 垂直方向に出せる速度がゼロに漸近する.
  そのためこのような脚構造では, 条件3(速度飽和の防止)を満たすためにある程度屈んだ状態で離陸する必要がある.
  この問題は回転関節脚であればGOAT\cite{Kalouche2017GOAT}のような非動物型の特殊なパラレルリンク構造の脚でも発生する.
  このように回転関節脚構造では条件2と条件3が相反しており, 高い跳躍の実現が阻害されている.

  一方でパラレルワイヤ脚構造には回転関節脚構造のような条件2と条件3の相反がなく, より簡単に3つの条件を満たすことができる.
  パラレルワイヤ脚構造では垂直方向の加速に直動関節を用いるため, 脚を伸ばしても垂直方向に出せる速度が下がらない.
  このため可動範囲を最大限使用して加速を行うことが可能である.
  さらにパラレルワイヤ脚構造は3つのモータを並列に用いて本体を持ち上げることができるため, 直列にアクチュエータを使用する脚構造\cite{boston2022atlas, bledt2018cheetah3}に比べて条件1を容易に満たせる.

  最後に連続跳躍を行う上で脚構造に求められる条件について確認する.
  連続跳躍のためには, 脚が軽いことが重要である.
  連続跳躍では脚を素早く動かしてロボットの姿勢を安定化させる.
  脚が本体に比べて重いと, 脚の最大加速度が低下し, さらに脚の反動で本体の姿勢が崩れやすくなる.
  パラレルワイヤ脚構造は, アクチュエータが一切搭載されていない軽量な脚を実現しており, 脚を俊敏に動かすことが可能である. 

  ここまでに高い跳躍と連続跳躍におけるパラレルワイヤ脚構造の利点を述べてきたが, パラレルワイヤ脚構造の短所として自重をアクチュエータの力のみで支える必要があることが挙げられる.
  動物のような回転関節の脚では膝を伸ばすことによって自重の大部分を関節で支えることが可能である.
  一方パラレルワイヤ脚では垂直方向に直動関節を使用しているため関節で自重を支えることができず, 立ち続けるだけで多くの電力を消費するという欠点がある.
  この短所は, 膝を伸ばしても垂直方向に出せる速度が低下しないという長所の裏返しであり, 消費電力を犠牲に跳躍に特化した結果であると考えられる.
 % 第一にパラレルワイヤ脚構造は本体を持ち上げるために, 3つのモータを並列に用いて大きな力を発揮することができる.
 % 本研究で開発したパラレルワイヤ一本脚ロボットRAMIELでは一つのモータが最大\SI{230}{\newton}の張力を発揮でき, ロボットの自重(\SI{10}{\kilogram})を考慮するとロボットは最大で重力加速度の6倍の加速度を発揮できる. 
  
 % 第二にパラレルワイヤ脚構造では垂直方向の加速に直動関節を使用するため, 回転関節を用いる脚構造に比べて加速距離を長く取ることができる.
 % 動物のように回転関節を垂直方向の加速に用いるロボット\cite{bledt2018cheetah3, boston2022atlas, kojima2019robot}では, 加速距離を増大させるためにリンク長を伸ばすことが求められる.
 % 回転関節脚でリンク長を伸ばす場合, 足先発揮力の維持のためにアクチュエータの出力トルクも同時に大きくする必要がある.
 % しかしアクチュエータの出力トルクを増大させると一般にアクチュエータの最大変位速度が低下するため, ロボットの跳躍高さが低下しうる. 
 % 一方で直動脚の場合は, 脚を延長して加速距離を増大させてもアクチュエータの出力トルクと足先発揮力の比は変化しない.
 % このためパラレルワイヤ脚構造は回転関節脚構造に比べて容易に加速距離を延長することができる.

%  第三にパラレルワイヤ脚構造は垂直方向の加速に直動関節を用いるため, 回転関節を用いる脚構造に比べて垂直方向の速度が加速中に飽和しづらい.
%  回転関節を垂直方向の加速に用いるロボット\cite{bledt2018cheetah3, boston2022atlas, kojima2019robot}は, 膝を伸ばしすぎると垂直方向の速度が飽和し跳躍できる高さが落ちてしまう.
%  言い換えると膝関節を一定角速度で開いて本体を持ち上げたとき, 膝が伸びれば伸びるほど本体が持ち上がる速度がゼロに漸近し, 跳躍高さが下がる.
%  このことから回転関節を持つロボットは垂直方向の速度の飽和を防ぐために, 膝をある程度曲げた状態で離陸する必要がある.
%  しかし膝を曲げた状態で離陸すると, 加速距離が縮まり跳躍高さが落ちてしまう. 
%  つまり垂直方向の加速に回転関節を用いる脚ロボットには, 加速距離と垂直方向の最大速度の間にトレードオフが存在する.
%  一方でパラレルワイヤは直動関節を採用しているため, 脚を伸ばしても垂直方向に出せる最大速度が下がらず, 加速距離を長くとっても垂直方向の速度飽和が発生しない.
  
  %このようにパラレルワイヤ脚構造は離陸時の速度を大きくする上で重要な, ロボットを持ち上げる力が大きい, 加速距離が大きい, 垂直方向の速度が加速中に飽和しづらいという3つの条件を満たしており高い跳躍に適していることが分かる. 

  %そして, ロボットが連続跳躍を行うためには脚が軽量であることが求められる.
  %脚が軽量だと, 脚の動作の反動がロボットの動作に与える影響を小さくすることができ, ロボットがバランスを保つことが容易になる.
  %また着地の際にロボットに加わる衝撃が小さくなり, 破損のリスクが低減される. 
  %パラレルワイヤ脚構造の脚部にはアクチュエータが搭載されておらず, 加速距離に対して脚部の質量を極めて小さくすることが可能である. 
  %実際RAMIELの脚部の質量は\SI{0.5}{\kilogram}となっており, これは本体質量\SI{10}{\kilogram}の\SI{5}{\percent}程度と極めて小さくなっている. 
}%

\begin{figure}[t]
  \centering
  \includegraphics[width=1.0\columnwidth]{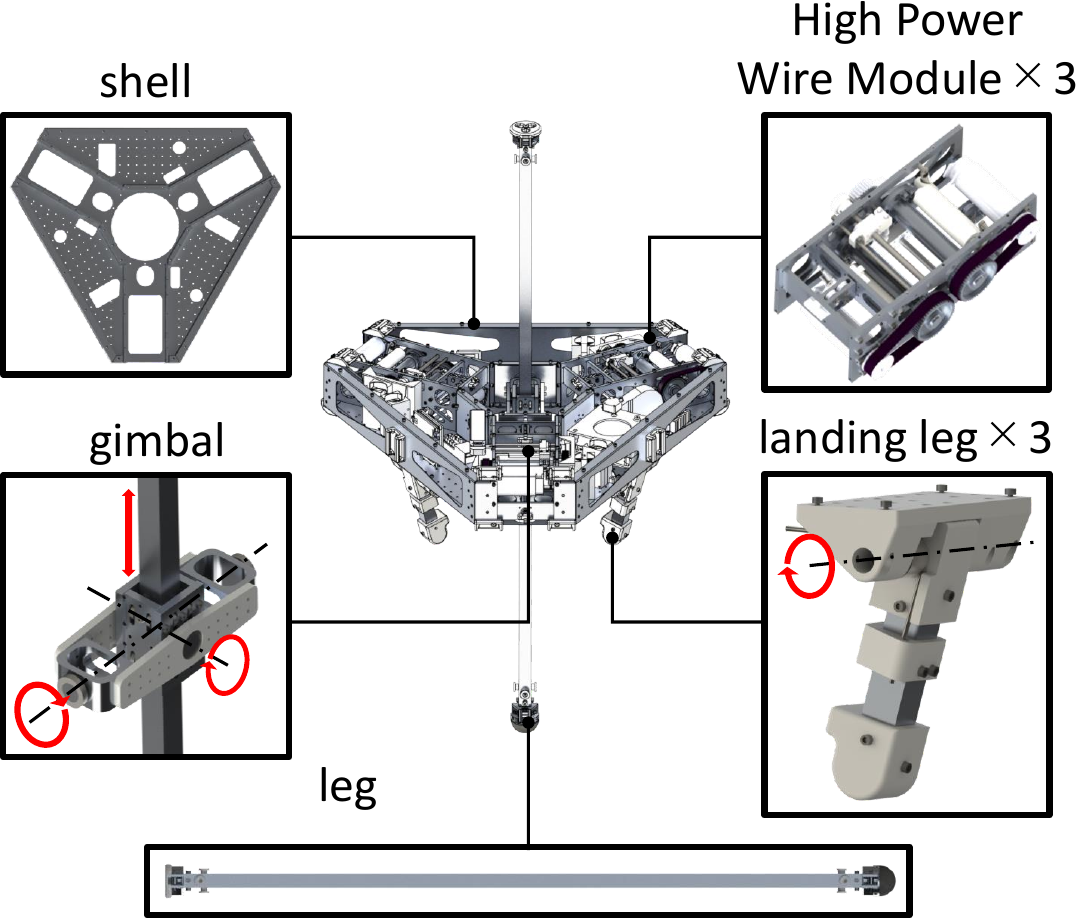}
  \vspace{-1.0ex}
  \caption{Detailed design of RAMIEL.}
  \label{figure:ramieldetail}
  %\vspace{3.0ex}
\end{figure}

\begin{table}[t]
  \begin{center}
  \caption{Physical parameters of RAMIEL.}
  \label{table:ramielparams}
  \begin{tabular}{lcr} \toprule
      Parameter & Value \\ \midrule
      Overall height  & \SI{1.07}{\metre} \\ 
      Overall width & \SI{0.55}{\metre} \\
      Total mass (body and leg) & \SI{10.3}{\kilogram} \\
      Body mass & \SI{9.8}{\kilogram} \\
      Leg mass  & \SI{0.5}{\kilogram} \\
      Body moment of inertia & \SI{0.225}{\kilogram.\metre^{2}} \\
      Leg moment of inertia & \SI{0.074}{\kilogram.\metre^{2}} \\
      Slide joint stroke  & \SI{0.8}{\metre} \\
      Roll joint movable range& \SI{-0.79}{\radian} to \SI{+0.79}{\radian}\\
      Pitch joint movable range& \SI{-0.79}{\radian} to \SI{+0.79}{\radian}\\
      Maximum wire tension& \SI{230}{\newton} at \SI{50}{A}\\
      Maximum force of leg slide joint  & \SI{690}{\newton}\\
      Maximum speed of leg slide joint at no load & \SI{15}{\metre/\s} at \SI{70}{\V}\\ \bottomrule
  \end{tabular}
  \end{center}
  \end{table}

\section{Design of Parallel Wire-driven Monopedal Robot} \label{sec:robot}
\subsection{Overview of Parallel Wire-driven Monopedal Robot} \label{subsec:designoverview}
\switchlanguage%
{%
%In this , \subsecref{subsec:designoverview} describes the overall design of RAMIEL, the parallel wire-driven monopedal robot developed in this study, and \subsecref{wiremodule} describes the detailed design of the wire module that controls the wire length. The detailed design of the wire module, which is controlled by subsecref{wiremodule}, is described.
To verify the jumping performance of the parallel wire-driven leg structure, we have designed and built a monopedal robot RAMIEL (pa{\bfseries RA}llel wire-driven {\bfseries M}onopedal ag{\bfseries I}l{\bfseries E} {\bfseries L}eg).
The details of RAMIEL are shown in \figref{figure:ramieldetail}. 
In addition, the specifications of RAMIEL are shown in \tabref{table:ramielparams}.
RAMIEL is a robot with a height of \SI{1.07}{\metre}, a width of \SI{0.55}{\metre}, and a weight of \SI{9.8}{\kilogram}. 
RAMIEL can be roughly divided into two parts: the leg and the body.
In order to increase the efficiency of the experiment, the power supply is installed outside. 
%The power supplies for the actuators and the control circuits are installed outside the robot to improve the efficiency of the experiments. 

In order to achieve high and continuous jumps, the leg has a lightweight structure consisting of only an aluminum square hollow pipe, wire attachment parts, and rubber for ground contact.
The leg has a simple configuration without actuators, which makes it light and controllable.

The body part consists of an outer shell as the frame, a gimbal module connecting the body and the leg, landing legs as grounding points when the body touches the ground, and High Power Wire Modules controlling the wires.
The detailed design of the wire module is described in the next section.
The outer shell is a monocoque structure made of aluminum with thicknesses of \SI{1}{\milli.\metre} and \SI{1.5}{\milli.\metre}, which is both lightweight and rigid.
In addition, cushions are installed on the surface of the outer shell to protect the contents when the side of the body part comes into contact with the ground.
The cushions are a simplified version of the pneumatic damper shock-absorbing outer shell \cite{takeda2019damper} developed by Takeda et al. 
In Takeda et al.'s method, a hole of about \SI{10}{\milli.\metre} in diameter was drilled in the surface of the cushion, and the damper function was realized by viscously releasing air through the hole when the cushion was impacted.
On the other hand, the cushions installed in RAMIEL have no air holes on its surface, and air leaks out through the seams of the fabric to realize the damper function more simply.

The gimbal has one DoF for rotation in the roll and pitch axes, respectively, and one Dof for linear motion parallel to pipe of the leg.
The linear motion mechanism is realized by pressing the square pipe of the leg from all sides with bearings attached to the inner wall of the gimbal. 
The gimbal does not have any actuators or sensors, and has a simple and robust structure.

The landing legs enable RAMIEL to start jumping from the ground and to return to the landing state after the completion of the jumping motion.
The landing leg has a rotational DoF at the base, and is almost vertical to the ground by the built-in torsion spring in the normal state.
When RAMIEL collides with the ground at an angle, the landing legs are folded from the base to absorb the impact. 
%This rotational DoF reduces the risk of damage when RAMIEL lands on the ground in a tilted position. 
}%
{%
  %この章では\subsecref{subsec:designoverview}で本研究で作成したパラレルワイヤ一本脚ロボットRAMIELの全体設計について述べ, \subsecref{wiremodule}でワイヤ長を制御するワイヤモジュールの詳細設計について述べる.
  パラレルワイヤ脚構造の跳躍性能を検証するため一本脚ロボットRAMIELを作成する. 
  なおRAMIELという名称はpa{\bfseries RA}llel wire-driven {\bfseries M}onopedal ag{\bfseries I}l{\bfseries E} {\bfseries L}egの略称である.

  RAMIELの詳細を\figref{figure:ramieldetail}に示す. 
  さらにRAMIELの諸元を\tabref{table:ramielparams}に示す。
  RAMIELは全高\SI{1.07}{\metre}, 全幅\SI{0.55}{\metre}, 重量\SI{9.8}{\kilogram}のロボットである. 
  RAMIELは大きく脚部分と本体部分に分けられる.
  なお実験効率を上げるために電源は外部に設置している. 

  高い跳躍と連続跳躍を達成するため, 脚部はアルミ角パイプ, ワイヤ固定パーツと接地用のゴムのみで構成された軽量な構造となっている.
  脚がアクチュエータのないシンプルな構成なため非常に軽量で制御性が高い.
  
  本体部分は, 骨格となる外殻, 本体-脚部をつなぐジンバルモジュール, 本体が接地する際に接地点となる着地脚, そしてワイヤを巻き取る大出力ワイヤモジュールから構成されている.
  なお, ワイヤモジュールの詳細設計については次節で述べる.
  外殻には厚さ\SI{1}{\milli.\metre}と\SI{1.5}{\milli.\metre}のアルミを用いたモノコック構造を採用しており, 軽量性と剛性を両立している.
  また外殻の表面にはクッションが搭載されており, 本体側面が地面と接触した際に内容物を保護している.
  クッションは武田らの開発した空気ダンパ衝撃吸収外装\cite{takeda2019damper}を簡素化したものである. 
  武田らの手法では布の表面に直径\SI{10}{\milli.\metre}程度の穴を開け, クッションに衝撃がかかった際にこの穴から空気が粘性を伴って抜けることでダンパ機能を実現していた.
  一方で今回RAMIELに搭載したクッションは表面に空気穴がなく, 布の縫い目から空気が漏れ出すことを利用してより簡易にダンパ機能を実現している.
  
  ジンバルはロール軸とピッチ軸にそれぞれ回転1自由度, 脚部の各パイプと平行に直動1自由度を持つ.
  直動機構は, 脚の角パイプ部分をジンバルの内壁に取り付けたベアリングで四方から押さえることで実現している. 
  ジンバルにはアクチュエータやセンサは搭載されておらず, 単純で頑健な構造である.
  
  着地脚は, RAMIEL本体が接地した状態から跳躍を開始し跳躍動作終了後に着地状態に復帰することを可能としている.
  着地脚は付け根に回転自由度が存在し, 通常時は内蔵されているねじりばねの力で地面に対して垂直に近い姿勢となっている.
  RAMIEL本体が地面に対して斜めに衝突した際には着地脚が付け根から折りたたまれることで衝撃を吸収する. 
  この回転自由度により, RAMIELが地面に対して傾いた状態で着地した際の破損のリスクを低減させることができる. 
}%

\subsection{Design of High Power Wire Module} \label{subsec:wiremodule}
\begin{figure}[t]
  \centering
  \includegraphics[width=1.0\columnwidth]{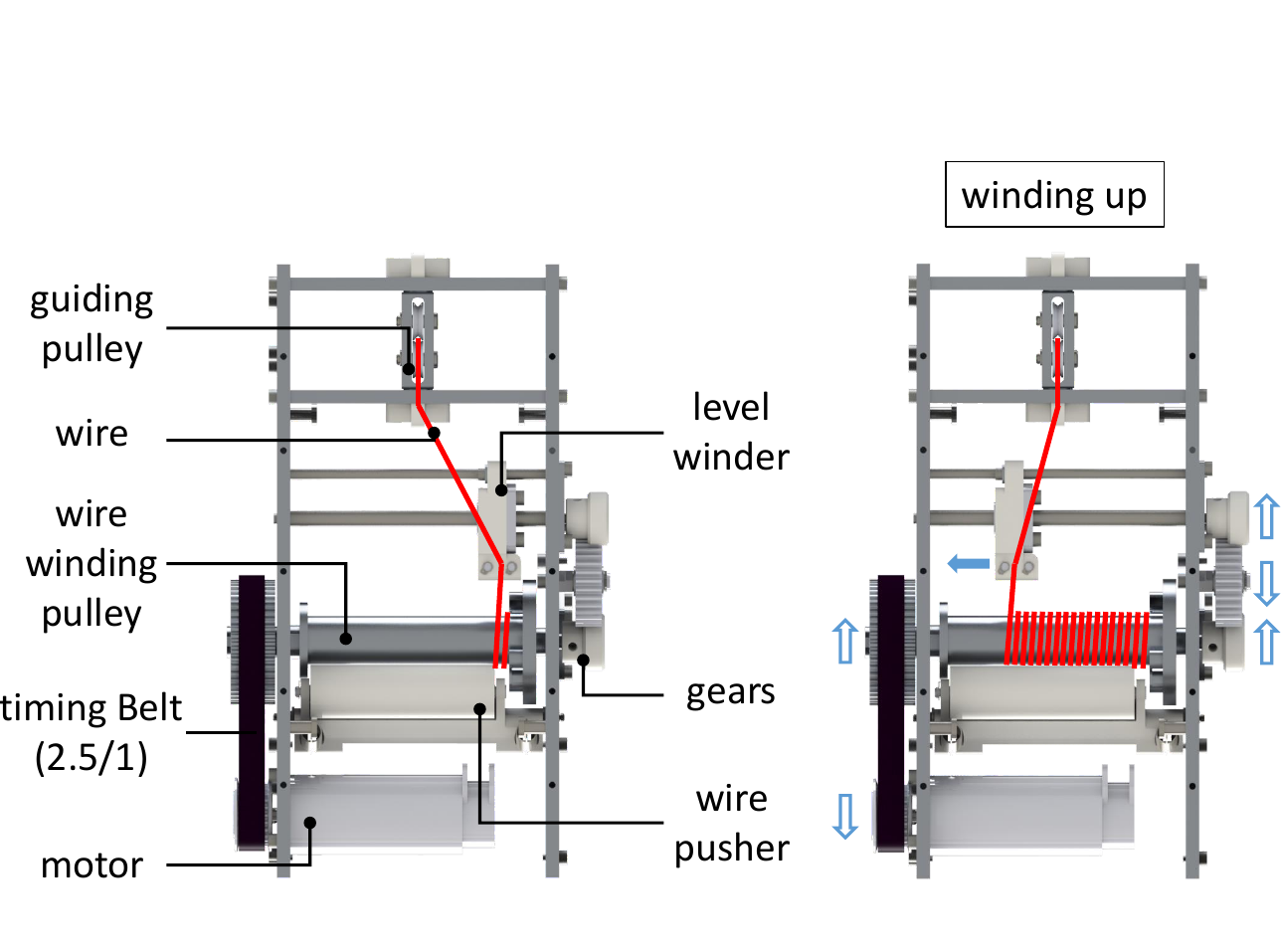}
  \vspace{-1.0ex}
  \caption{Detailed design of High Power Wire Module.}
  \label{figure:wiremodule}
  \vspace{-3.0ex}
\end{figure}

\switchlanguage%
{%
  RAMIEL uses six wires to control the posture of the leg and is equipped with three sets of modules to control the length of the wires. 
  One wire module controls the winding length of two wires independently.  
  The details of the High Power Wire Module installed in RAMIEL are shown in \figref{figure:wiremodule}. 
  The wire module must satisfy the following three conditions to achieve high and continuous jumps.
  \begin{enumerate}
    \item Tension and maximum velocity are large enough for jumping.
    \item Will not be destroyed by the impact of landing.
    \item Be able to wind the wire exactly to the intended length.
  \end{enumerate}
  In this study, a High Power Wire Module is developed to meet these three requirements.
  In the following, we review these three requirements in detail and explain how the wire module satisfies them.

  First, both the maximum winding speed and the maximum tension of the wire must be sufficiently high to achieve a high jump. 
  The High Power Wire Module satisfies this requirement by operating BLDC motors (Maxon \cite{maxon2018maxon}) that winds the wire under the high-power conditions of input voltage \SI{70}{\volt} and maximum instantaneous current \SI{50}{\ampere} \cite{sugai2018design}. 
  The output shaft of the motor is reduced to 2.5/1 by a timing pulley and input to a wire winding pulley with a diameter of \SI{20}{\milli.\metre}. 
  The maximum tension of the wires is about \SI{230}{\newton} (\SI{50}{\ampere} is applied), and approximating the parallel wire-driven leg structure as a simple linear motion, the three wires can lift the robot body with a total force of \SI{690}{\newton}.
  This means that RAMIEL can be accelerated at six times the gravitional acceleration.
  In addition, the high input voltage of the motor, \SI{70}{\volt}, allows the maximum wire winding speed to be as fast as \SI{10.7}{\metre/\sec} even with a high current of \SI{50}{\ampere}.
  Assuming that the parallel wire-driven leg structure is approximated as a simple linear motion and the robot is ejected vertically upward at \SI{10.7}{\metre/\sec}, the robot can theoretically jump to a height of about \SI{5.8}{\metre}.

  Secondly, a major problem for a jumping robot is that a large impact on landing can destroy the leg and actuators. 
  In the High Power Wire Module, the reduction ratio of the motor is set to be as low as 2.5/1 and the quasi-direct drive is adopted to solve this problem.
  %The quasi-direct drive reduces the impact on the leg and actuators due to the backdrive of the wire at the time of landing.
  The quasi-direct drive causes the wire to backdrive when landing, reducing the impact on the leg and actuators.

  Thirdly, for accurate posture control of the leg in continuous jumping, it is necessary to wind the wire precisely for the intended length. 
  In the High Power Wire Module, a level winder and a wire pusher are used to achieve accurate wire winding. 
  The wire from the wire winding pulley passes through the level winder to the guiding pulley.
  The level winder moves linearly on the sliding screw in conjunction with the rotation of the wire winding pulley, and plays the role of aligning the wire on the winding pulley.
  The wire puhser prevents the wire from crossing each other on the winding pulley and assists smooth winding.
  The wire has a diameter of \SI{2}{\milli.\metre} with a zylon{\textregistered} core yarn and polyester side yarns, and the wire elongation is considered to have a small effect on the winding length because the wire elongation is only about \SI{1}{\percent} even when the maximum tension of the wire module of \SI{230}{\newton} is applied. 
}%
{%
  RAMIELは6本のワイヤを用いて脚の姿勢を制御しており, このワイヤ長さを制御するモジュールが3組搭載されている. 
  1つのワイヤモジュールは2本のワイヤの巻取り長さをそれぞれ独立に制御している.  Therefore, the effect of wire elongation on the winding length is considered to be small.
  RAMIELに搭載されている大出力ワイヤモジュールの詳細を\figref{figure:wiremodule}に示す. 
  ワイヤモジュールは高い跳躍と連続跳躍を実現するために以下の3つの条件を満たす必要がある.
  \begin{enumerate}
    \item 張力と最大速度が, 跳躍に十分なほど大きい
    \item 着地の衝撃で破壊されない
    \item ワイヤを意図した長さ正確に巻き取れる
  \end{enumerate}
  本研究ではこれら3つの要件を満たすべく, 大出力ワイヤモジュールを新規に開発した.
  以下これら3つの要件について詳細に確認し, ワイヤモジュールがこの要件をどのように満たしているのかを説明する.

  第一に, 高い跳躍を実現するためにはワイヤの最大巻取り速度とワイヤの最大張力が両方とも十分大きいことが求められる. 
  大出力ワイヤモジュールは, 菅井らのモータドライバ\cite{sugai2018design}を用いてワイヤを巻き取るBLDC モータ(Maxon \cite{maxon2018maxon})を入力電圧\SI{70}{\volt}, 最大瞬間電流\SI{50}{\ampere}という大出力な条件で運用することでこの要求を満たしている. 
  なおモータの出力軸はタイミングプーリによって2.5/1に減速されて直径\SI{20}{\milli.\metre}のワイヤ巻取りプーリに入力される. 
  ワイヤの最大張力は約\SI{230}{\newton}(\SI{50}{\ampere}を印加)であり, パラレルワイヤを単純な直動に近似すると3つのワイヤで合計\SI{690}{\newton}の力でロボット本体を持ち上げることができる.
  これは, \secref{sec:parallel-wire-driven-leg}で述べた通りRAMIELを重力加速度の6倍で加速させられることを意味する.
  さらにモータの入力電圧が\SI{70}{\volt}と高いことで, \SI{50}{\ampere}の大電流を印加した状態でも最大ワイヤ巻取り速度が\SI{10.7}{\metre/\sec}と高速になっている.
  パラレルワイヤを単純な直動に近似しロボットが垂直上向きに\SI{10.7}{\metre/\sec}で射出されたと仮定すると, ロボットは約\SI{5.8}{\metre}の高さまで跳躍することができる.

  第二に, 跳躍ロボットでは着地の際に大きな衝撃がかかり脚やアクチュエータが破壊されることが大きな問題となる. 
  大出力ワイヤモジュールではモータの減速比が2.5/1と低く設定し準ダイレクトドライブにすることでこの問題に対応している.
  準ダイレクトドライブにより, 着地の際にワイヤがバックドライブし, 脚やアクチュエータに加わる衝撃が小さくなっている.

  第三に, 連続跳躍における脚の正確な姿勢制御のためには, 意図した長さの分正確にワイヤを巻き取る必要がある. 
  大出力ワイヤモジュールでは, レベルワインダとワイヤ押さえを用いてワイヤの正確な巻取りを実現している. 
  ワイヤ巻取りプーリからでたワイヤは, レベルワインダ内部を通過してワイヤ経由プーリへと向かう.
  レベルワインダはワイヤ巻取りプーリの回転に連動して滑りねじ上を直動し, ワイヤ巻取りプーリ上でワイヤを整列させる役割を果たしている.
  ワイヤ押さえはワイヤが巻取りプーリ上で交差することを防ぎ, 円滑な巻取りを補助している.
  なおワイヤには直径\SI{2}{\milli.\metre}で芯糸がzylon \textregistered , 側糸がポリエステルのものを使用しており, ワイヤモジュールの最大張力\SI{230}{\newton}を印加しても約\SI{1}{\percent}しか伸びないためワイヤの伸びが巻取り長さに与える影響は小さいと考えられる.
}%

\section{Controller of Parallel Wire-driven Monopedal Robot} \label{sec:controller}

\begin{figure*}[t]
  \centering
  % https://drive.google.com/file/d/13pg6vb1YBGpngBPxdnkizgbrhkr0AKY2/view?usp=sharing
  \includegraphics[width=2.0\columnwidth]{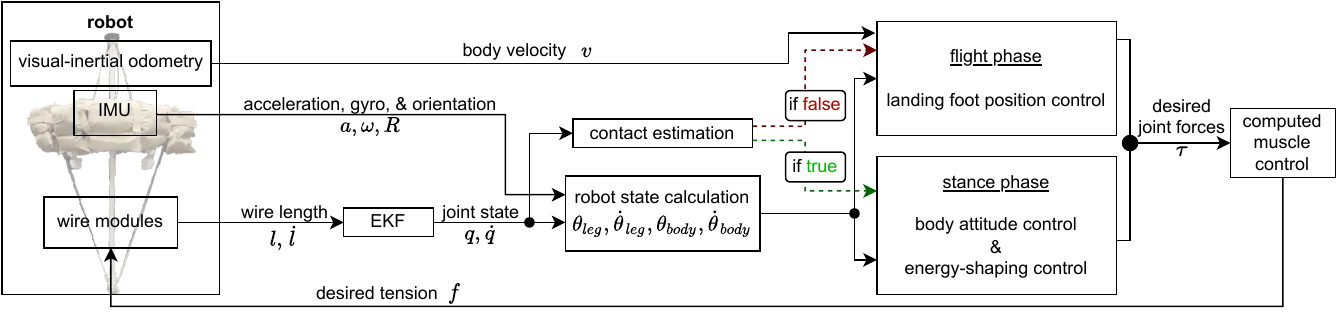}
  \caption{Overview of the hopping controller.}
  \label{figure:hopcontrol_system_diagram}
\end{figure*}

% スペースが厳しいならこの図はなくてもいいかも
\begin{figure}[tbh]
  \begin{minipage}{0.6\hsize}
      \centering
      \includegraphics[width=\textwidth]{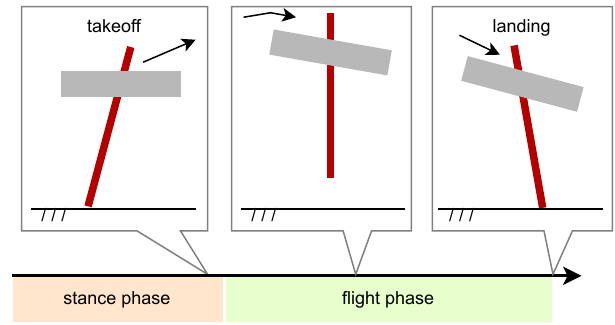}
      \caption{Phases of hopping.}
      \label{figure:phases}
    \end{minipage}
  \begin{minipage}{0.39\hsize}
    \centering
    \includegraphics[width=0.95\columnwidth]{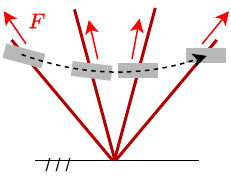}
    \caption{Force encountered during stance phase.}
    \label{figure:stance_force}
  \end{minipage}
\end{figure}

\switchlanguage%
{% 
  In this section, we describe the controller for RAMIEL's hopping operation, which is based on the controller of the 3D One-Legged Hopping Machine of Raibert et al. with modifications to support a parallel wire-driven leg strcture \cite{marc1984onelegged}.
  
  \figref{figure:hopcontrol_system_diagram} shows an overview of the controller. The controller assumes that the motion in the pitch and roll directions (in the XZ and YZ planes, respectively) are decoupled from each other, so the same processing is performed for each direction.
  Therefore, for simplicity, the following explanation is limited to the pitch direction, i.e., the XZ plane.

  The inputs of the controller are the wire length $l$ and velocity $\dot l$ obtained from the encoders of each motor, the acceleration $a$ and angular velocity $\omega$ obtained from the IMU fixed to the body part, and the estimated velocity $v$ of the body part as a result of visual-inertial odometry performed by the RealSense T265. In addition, a Magdwick filter is used to estimate the rotational posture $R$ from the IMU sensor information \cite{madgwick2010efficient}. Since the joint does not have a built-in encoder, the joint posture is estimated from the wire length and velocity using the Extended Kalman Filter (EKF).
  
  The EKF is constructed as follows. First, the state $\bm x$ to be estimated is defined from the joint position $\bm q$ and the joint velocity $\dot{\bm q}$, and the observed value $\bm z$ is defined as a vector consisting of the wire length $\bm l$ and the wire velocity $\dot{\bm l}$ as follows.
  \begin{equation}
    \bm x = 
    \begin{bmatrix}
        \bm q \\ \bm{\dot q}  
    \end{bmatrix}
    \in \mathbb R^{6}
    \quad
    \bm z = 
    \begin{bmatrix}
        \bm l \\ \dot{\bm l}
    \end{bmatrix}
    \in \mathbb R^{12}
  \end{equation}
   The state transition model can then be defined as follows.
   \begin{equation}
    \bm x_{i+1}
    =
    \begin{bmatrix}
    I && I dt \\
    O && I
    \end{bmatrix}
    \bm x_i
    + \bm w
  \end{equation}
  where the subscript $i$ denotes the data at step $i$, $I$ is the unit matrix, $O$ is the zero matrix, $dt$ is the time step of the state transition, and $\bm w$ is the noise of the model.
  The observation model can be defined as follows.
  \begin{equation}
    \begin{split}
        \bm z_i
        &=
        \bm h(\bm x_i)
        + \bm v
        \\
        \bm h(\bm x_i) &:= \begin{bmatrix}
            \bm g(\bm q_i)
            \\
            J_m(\bm q_i) \dot{\bm q}_i
        \end{bmatrix}
    \end{split}
  \end{equation}
  The model is nonlinear. Here, $\bm g(\cdot)$ and $J_m(\cdot)$ are functions that calculate the wire length and muscle jacobian from the joint position, respectively. $\bm g(\cdot)$ is a function that calculates the wire path by geometric calculation based on the dimensional information of the CAD model, and its analytical partial differentiation is used as $J_m(\cdot)$\cite{autodiff}. The lower wires tended to become slack upon landing, as they are quickly shortened while carrying little tension. This results in an error of the joint angle output from the EKF, and to prevent this the measurements of the lower wires were weighted to have less effect on the estimation for \SI{0.1}{s} after touchdown was detected.

  In the control of the hopping motion, the joint force $\bm \tau \in \mathbb R^3$, which is a combination of the torque of the rotary joint and the force of the linear joint, is output, and it is converted to the wire tension $\bm f$ before being sent to the wire module of the robot. The conversion from $\bm \tau$ to $\bm f$ is done using computed muscle control, which can be formulated as a quadratic programming (QP) method using the muscle jacobian $J_m$ as follows \cite{jantsch2013anthrob, kawamura2016jointspace}.
  \begin{equation}
    \label{eq:computed-torque-control}
        \begin{split}
            \argmin_{\bm f}& \quad || \bm f || ^2
            \\
            \text{subject to}&
            \begin{cases}
                & \bm \tau = - J_m^T \bm f
                \\
                &\bm f \geq \bm f_{min}
            \end{cases}
        \end{split}
  \end{equation}  
  $||\cdot||^2$ denotes the L2 norm, and $\bm f_{min}$ denotes the minimum tension required to keep the wire taut.

  As shown in \figref{figure:phases}, the hopping motion can be divided into a stance phase, in which the feet are touching the ground, and a flight phase, in which the feet are away from the ground. At the moment of transition to the stance phase, i.e., at the time of landing, the robot has a vertical downward velocity component and a horizontal velocity component. After that, as shown in \figref{figure:stance_force}, the robot sinks and rises like a spring throughout the stance phase, and enters the flight phase when the normal force against the ground becomes zero. In the flight phase, the motion changes with an acceleration of $-g$ in the vertical direction.

  In the stance phase, PD control is applied to the rotational joints so that the posture of the body is horizontal. For the linear motion direction, the summed force of a constant feed-forward element $F_{F}$ and a feedback element $F_{B}$ are generated to accelerate the body upward.
  The feedback element $F_{B}$ is determined by the energy shaping control law as shown in the following equation, depending on the difference between the target and actual mechanical energy to maintain an appropriate hopping height.
  \begin{equation}
    \begin{split}
      \label{eq:energy_fb}
      F_{B}(x, \dot x) &= k_E (E_{des} - E(x, \dot x)) \dot x
      \\
      E_{des} &:= \frac{1}{2} m_B \dot{x}_{des}^2
      \\
      E(x, \dot x) &:= \frac{1}{2} m_B \dot x ^ 2 +  F_{F} x - m_B g (x - x_0)
    \end{split}
    \end{equation}
    Here, $x$ is the linear joint position, $k_E$ is the feedback gain, $m_B$ is the mass of the body, and $g$ is the gravitational acceleration. The $E_{des}$ is the target value of the mechanical energy, which in the above equation is determined from the vertical velocity $\dot x_{des}$ during takeoff, but it can also be determined from the target hopping height.

  During the flight phase, the robot should maintain a constant position $x_0$ for the linear joint so that the length of the leg does not change. In the rotational direction, based on the horizontal velocity $v_H$ of the robot, the target foot position $p_0$ is determined by the following feedback law, which is a simplified version of the control law \cite{marc1984onelegged} of Raibert et al., to keep it hopping in place.
  \begin{equation}
    p_0 = k v_H
  \end{equation}
  Here, $k$ is the gain, and after obtaining a stable value on the simulator, we manually adjusted it to be stable on the actual machine.
  The horizontal velocity $v_H$ was obtained from the visual inertial odometry output from the RealSense T265 attached to the robot body.

  The controller judges that it is the flight phase if the position of the linear joint is closer to $x_0$ than the threshold value, and the stance phase otherwise.  

  Finally, the process at the beginning and end of the hopping motion is described.
  At the start of the hopping motion, the robot is grounded on the landing leg as shown in \figref{figure:experimentssetup}. Then, after applying a force to accelerate the robot upward in the linear direction for a certain period of time, the ascent speed is adjusted by the feedback law of \equref{eq:energy_fb} until the transition to the stance phase, and the robot jumps up at the target speed. After completing the hopping motion, the robot returns to ground contact with the landing leg, which was achieved by setting a large D gain to decelerate against the linear motion direction.
}%
{%
  本節では，RAMIELのホッピング動作を実現するための制御器を解説する．Raibertらの3D One-Leg Hopperの制御器を踏襲しており，パラレルワイヤ駆動方式に対応するために変更を加えている\cite{marc1984onelegged}．
  
  \figref{figure:hopcontrol_system_diagram}に制御器の概要を示す．なお，制御器ではピッチ方向とロール方向の動き(それぞれ，XZ平面とYZ平面内の動き)は互いに影響しない(decoupled)と仮定しているため，それぞれの方向について同じ処理を行なっている．
  そのため，以降の説明では簡単のためピッチ方向のみ，つまりXZ平面に限定して説明する．

  制御に利用される入力は各モーターのエンコーダから得られるワイヤ長$l$と速度$\dot l$，本体に固定されているIMUから得られる加速度$a$，角速度$\omega$，およびRealSense T265で行われるVisual-inertial odometryの結果として出力される本体の推定速度情報$v$である．また，Magdwickフィルターを用いてIMUのセンサ情報から回転姿勢$R$を推定する\cite{madgwick2010efficient}．関節にはエンコーダが内蔵されていないため，関節姿勢はワイヤ長・速度から，拡張カルマンフィルタ(EKF)を用いて推定する．

  % EKFのくだりはスペース足りなかったら大幅に削っても大丈夫

  EKFは次のように構成される．まず，推定対象の状態$\bm x$は関節位置$\bm q$と関節速度$\dot{\bm q}$から，そして観測値$\bm z$はワイヤ長$\bm l$とワイヤ速度$\dot{\bm l}$から構成されるベクトルとして以下のように定義する．
  \begin{equation}
    \bm x = 
    \begin{bmatrix}
        \bm q \\ \bm{\dot q}  
    \end{bmatrix}
    \in \mathbb R^{6}
    \quad
    \bm z = 
    \begin{bmatrix}
        \bm l \\ \dot{\bm l}
    \end{bmatrix}
    \in \mathbb R^{12}
  \end{equation}
  すると状態遷移モデルは以下のように定義できる．
  \begin{equation}
    \bm x_{i+1}
    =
    \begin{bmatrix}
    I && I dt \\
    O && I
    \end{bmatrix}
    \bm x_i
    + \bm w
  \end{equation}
  ここで，下付き文字$i$はステップ$i$におけるデータを示し，$I$は単位行列，$O$は零行列を表し，$dt$は状態遷移の時間ステップ，$\bm w$はモデルのノイズを表す．
  観測モデルは以下のように定義できる．
  \begin{equation}
    \begin{split}
        \bm z_i
        &=
        \bm h(\bm x_i)
        + \bm v
        \\
        \bm h(\bm x_i) &:= \begin{bmatrix}
            \bm g(\bm q_i)
            \\
            J_m(\bm q_i) \dot{\bm q}_i
        \end{bmatrix}
    \end{split}
  \end{equation}
  という非線形なモデルとなる．ここで，$\bm g(\cdot)$と$J_m(\cdot)$はそれぞれ関節位置姿勢からワイヤ長と筋長ヤコビアンを求める関数である．$\bm g(\cdot)$はCADモデルの寸法情報から幾何学計算によりワイヤ経路を計算した関数であり，それを解析的に偏微分したものを$J_m(\cdot)$として用いた\cite{autodiff}．

  また，ホッピング動作の制御では回転関節のトルクと直動関節の力を合わせた関節力$\bm \tau \in \mathbb R^3$が出力されるので，それをワイヤ張力$\bm f$に変換してからロボットのワイヤモジュールに送る．$\bm \tau$から$\bm f$の変換は，筋長ヤコビアンを利用して以下のような二次計画法(QP)として定式化できるcomputed muscle controlを利用した\cite{jantsch2013anthrob, kawamura2016jointspace}
  \begin{equation}
    \label{eq:computed-torque-control}
        \begin{split}
            \argmin_{\bm f}& \quad || \bm f || ^2
            \\
            \text{subject to}&
            \begin{cases}
                & \bm \tau = - J_m^T \bm f
                \\
                &\bm f \geq \bm f_{min}
            \end{cases}
        \end{split}
  \end{equation}  
  $||\cdot||^2$はL2ノルムを示し，$\bm f_{min}$はワイヤが張った状態を維持するために必要な最低張力を表す．

  ホッピング動作は，\figref{figure:phases}に示すように，足が地面に触れている立脚期(stance phase)と離れている滞空期(flight phase)に分けられる．立脚期に移行する瞬間，つまり着陸時にロボットは鉛直下向きの速度成分と水平方向の速度成分を持つ．その後，立脚期を通して\figref{figure:stance_force}に示したとおりバネのように沈み込んでから再び上がっていく運動が発生し，地面に対する垂直抗力がゼロになったタイミングで，滞空期に移行する．滞空期では鉛直方向に$-g$の加速度で動きが変化していく．

  立脚期には，回転関節に対しては本体の姿勢が水平となるようなPD制御を行う．直動方向に対しては，本体を上向きに加速させる方向に働く一定のフィードフォワード要素$F_{F}$と，フィードバック要素$F_{B}$の合計$F_{F} + F_{B}$を発生させる．
  フィードバック要素$F_{B}$は，適切なホッピング高さを維持するために目標と実際の力学的エネルギーの差に応じて，以下の式のようなenergy shaping control則により決定される．
  \begin{equation}
  \begin{split}
    \label{eq:energy_fb}
    F_{B}(x, \dot x) &= k_E (E_{des} - E(x, \dot x)) \dot x
    \\
    E_{des} &:= \frac{1}{2} m_B \dot{x}_{des}^2
    \\
    E(x, \dot x) &:= \frac{1}{2} m_B \dot x ^ 2 +  F_{F} x - m_B g (x - x_0)
  \end{split}
  \end{equation}
  ここで，$x$は直動関節位置，$k_E$はフィードバックゲイン，$m_B$は本体の質量，$g$は重力加速度を表す．$E_{des}$は力学的エネルギーの目標値で，上式では離陸時の垂直方向の速度$\dot x_{des}$から決定しているが，目標のホッピング高さなどから決定することもできる．

  滞空期には，直動方向については一定位置$x_0$を保ち，足の長さが変化しないようにする．回転方向については，ロボットの水平方向の速度$v_H$を元に，Raibertらの制御則\cite{marc1984onelegged}を簡略化した以下のフィードバック則により足先位置の目標値$p_0$を決定し，それを達成するようにPD制御を行う．
  \begin{equation}
    p_0 = k v_H
  \end{equation}
  ここで$k$はゲインであり，シミュレータ上で安定する値を求めた後に実機でも安定するよう調整を行った．また，水平方向速度$v_H$はロボット本体に取り付けられたRealSense T265から出力されるvisual inertial odometryの結果を利用した．

  なお，この制御器では直動関節の位置が$x_0$に閾値よりも近ければ滞空期，遠ければ立脚期と判断している.

  最後に，ホッピング動作の開始時と終了時の処理について述べる．
  ホッピングの開始時点でロボットは\figref{figure:experimentssetup}のように着地脚で接地している状態である．その後，直動方向にロボットを上向きに加速させる力を一定時間かけた後，滞空期に移行するまで\equref{eq:energy_fb}のフィードバック則により上昇速度を調整し，目標速度で飛び上がる．また，ホッピング動作を終了してからも着地脚で接地する状態に戻るが，これは直動方向に対して減速するために大きなDゲインを設定することで達成された．
}%

\section{Experiments} \label{sec:experiments}
\subsection{Experimental Condition} \label{subsec:experimental-condition}
\begin{figure}[t]
  \centering
  \includegraphics[width=0.6\columnwidth]{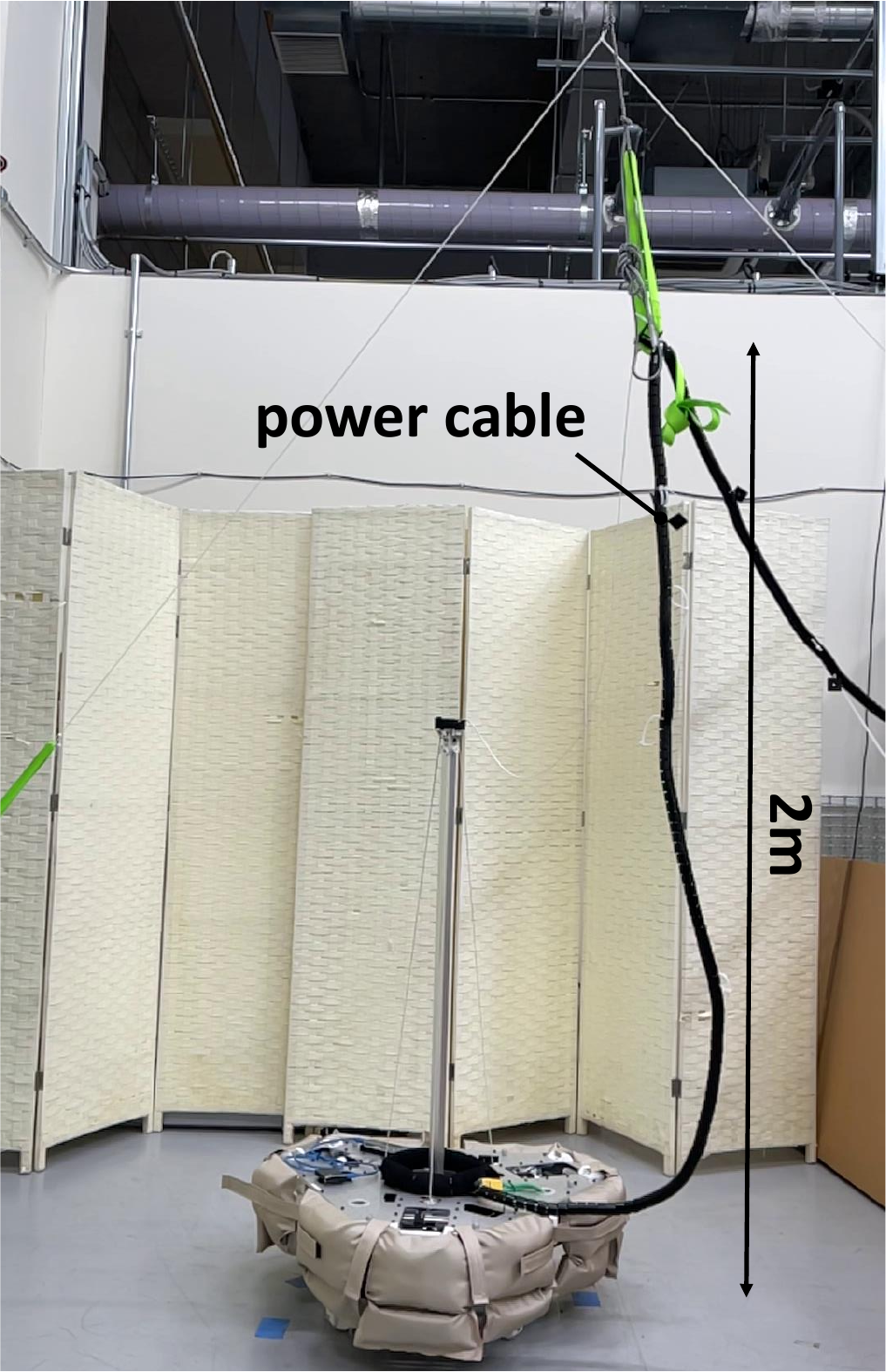}
  \vspace{-1.0ex}
  \caption{Initial state of RAMIEL in the jumping experiments.}
  \label{figure:experimentssetup}
  \vspace{-3.0ex}
\end{figure}

\switchlanguage%
{%
In this chapter, we perform high jump experiments and continuous jump experiments using RAMIEL, and discuss the results of each experiment.
The experimental setup is shown in \figref{figure:experimentssetup}.
All experiments are performed on a horizontal floor. 
The power to drive the motor and the control circuit of RAMIEL are both supplied from outside through cables. 
The power cable extending from RAMIEL is suspended from a height of \SI{2}{\metre}. 

As mentioned earlier, RAMIEL is not equipped with an encoder to read the joint angle directly, and the joint angle is estimated from the displacement of the wire length.
The initial value for joint angle estimation was set as follows. 
First, the joint is fixed to a known joint angle using a jig. 
Next, the wire is wound with a small tension of \SI{0.5}{\newton}.
The initial length is set to the wire winding length when the wire is no longer loose.
During the jump, the difference between the initial length and the current wire length is used to estimate the current joint angle.
}%
{%
  本章ではRAMIELによる高い跳躍実験と連続跳躍実験を行い, それぞれの結果について考察する.
  実験のセットアップを\figref{figure:experimentssetup}に示す。
  実験はすべて水平な床面上で行う. 
  RAMIELのモータを駆動する電源と制御回路用電源はどちらもケーブルを通して外部から供給されている. 
  RAMIELから伸びる電源ケーブルは\SI{2}{\metre}の高さから吊り下げられている. 

  前述の通りRAMIELには関節角度を直接読み取るエンコーダが搭載されておらず, ワイヤ長の変位から関節角度を推定している.
  関節角度推定の初期値は次のようにして設定した. 
  まず治具を用いて関節を既知の関節角度へ固定する. 
  次にワイヤを\SI{0.5}{\newton}程度の小さい張力で巻取る.
  そしてワイヤの緩みが無くなったときのワイヤ巻取り長さを, 初期長に設定する.
  跳躍中は, 初期長と現在のワイヤ長の差分を用いて現在の関節角度を推定する.
}%

\subsection{High Jump} \label{subsec:high-jump}
\begin{figure}[t]
  \begin{center}
   \includegraphics[width=0.9\columnwidth]{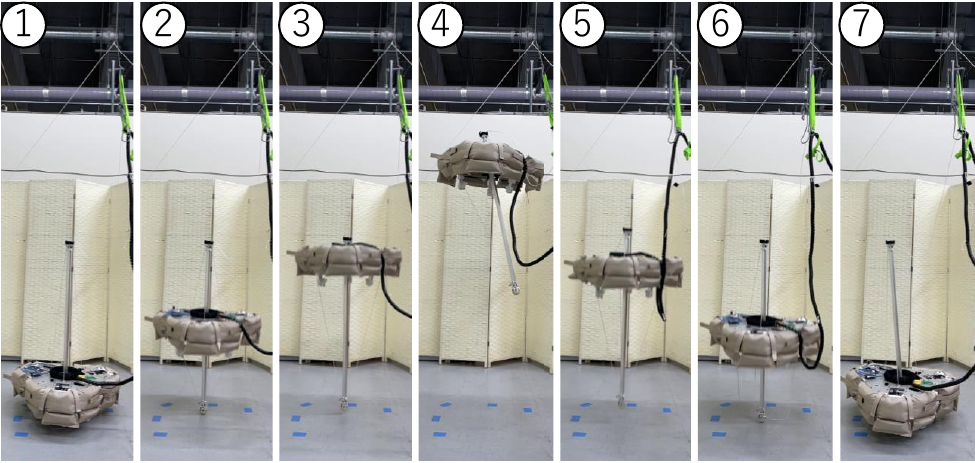}
   \caption{Snapshots of \SI{0.7}{\metre} high jumping motion. RAMIEL can jump and land from seated position.}
   \label{figure:20220114onehop1mpicture}
  \end{center}
\end{figure}

\begin{figure}[t]
  \begin{center}
   \includegraphics[width=0.8\columnwidth]{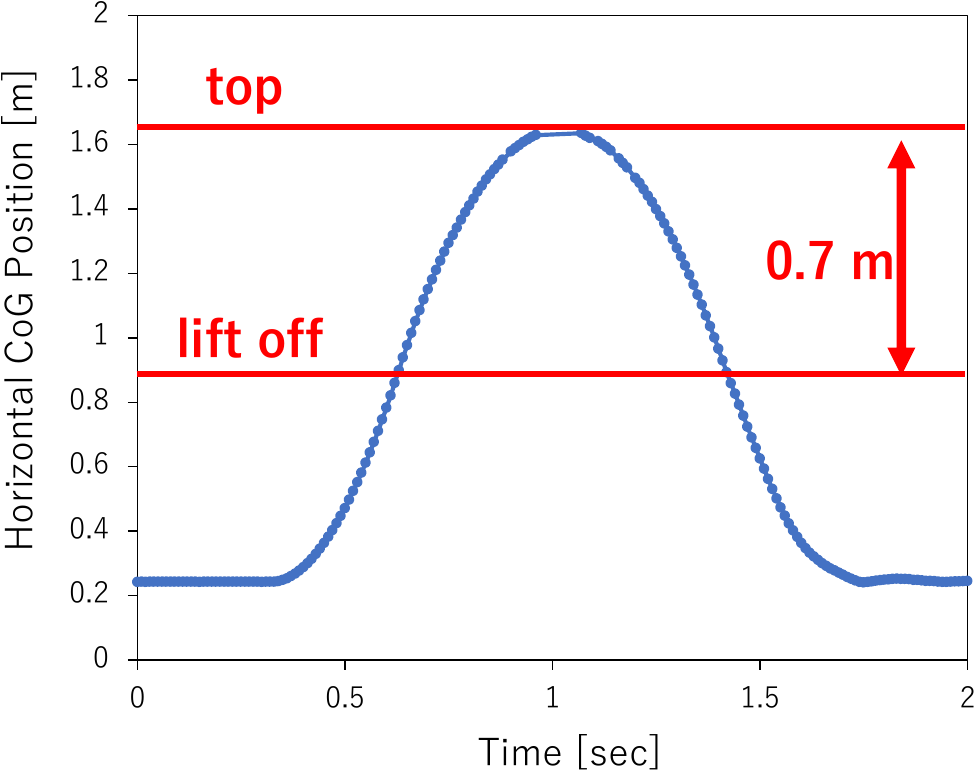}
   \caption{Vertical position of RAMIEL's center of gravity (CoG) during a jump of \SI{0.7}{m} in height.}
   \label{figure:20220114onehop1mtz}
  \end{center}
\end{figure}

\begin{figure}[t]
  \begin{center}
   \includegraphics[width=1.0\columnwidth]{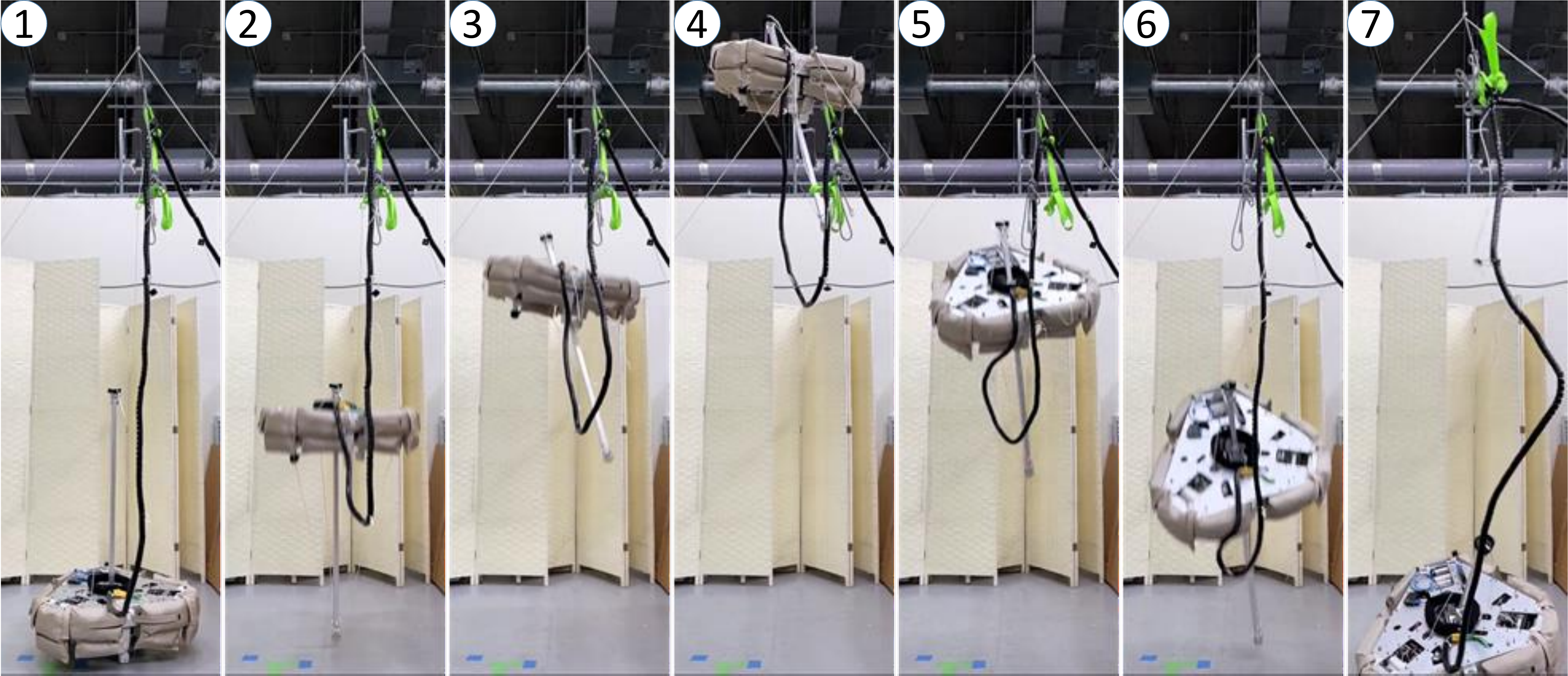}
   \caption{Snapshots of \SI{1.6}{\metre} high jumping motion. RAMIEL fails to land.}
   \label{figure:20220201onehop2mpicture}
  \end{center}
\end{figure}

\begin{figure}[t]
  \begin{center}
   \includegraphics[width=0.8\columnwidth]{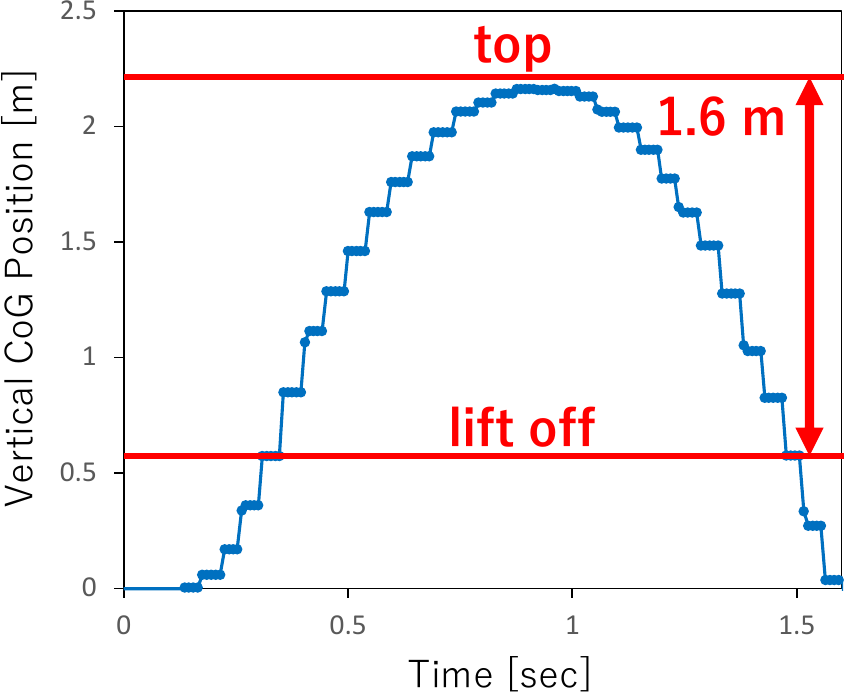}
   \caption{Vertical position of RAMIEL's center of gravity (CoG) during a jump of \SI{1.6}{\metre} in height.}
   \label{figure:20220201onehop2mtz}
  \end{center}
\end{figure}

\switchlanguage%
{%
  First, we perform a jumping experiment with a commanded CoG jump height of \SI{1}{\metre}. 
  RAMIEL during the jump is shown in \figref{figure:20220114onehop1mpicture}.
  It can be seen that the body of RAMIEL jumps from the horizontal ground position and returns to the starting position after the jump.
  The vertical displacement of the CoG of RAMIEL is shown in \figref{figure:20220114onehop1mtz}.
  The height of the CoG of RAMIEL at takeoff is \SI{0.9}{\metre}, and the maximum height is \SI{1.63}{\metre}, indicating that the CoG of RAMIEL is displaced vertically by a maximum of \SI{0.73}{\m}.
  The vertical displacement of the CoG in the jump experiment with the commanded jump height of \SI{1}{\metre} was measured using motion capture. 

  Next, we will perform a jumping experiment with the commanded jump height of the CoG \SI{2}{\metre}. 
  RAMIEL during the jump is shown in \figref{figure:20220201onehop2mpicture}.
  It can be seen that the jump is successful, but the landing is not. 
  In addition, while the robot is in the air, one of the upper wires snapped and the other wires became loose. 
  The vertical displacement of the CoG of RAMIEL is shown in \figref{figure:20220201onehop2mtz}.
  The vertical displacement of the CoG is measured using an RealSense T265 attached to RAMIEL body, because RAMIEL exceed the view of the motion capture system in the jumping experiment of the commanded CoG jump height \SI{2}{\metre}.
  Since the CoG is measured using RealSense, there is an offset in the height of the CoG compared to that using motion capture.
  Since the CoG jump height is the difference between the CoG height at the moment when the foot leaves the ground and the CoG height at the heighest point, the same CoG jump height is measured by both RealSense and motion capture regardless of the offset of the CoG height. Therefore, the same CoG jump height is measured by both RealSense and motion capture regardless of the offset.
  The height of the CoG of RAMIEL at takeoff was \SI{0.57}{\metre}, and the maximum height was \SI{2.16}{\metre}, indicating that the CoG of RAMIEL was displaced vertically by a maximum of \SI{1.59}{\m}.
  
  We discuss the results of these experiments.
  The experimental results for the commanded CoG jump height \SI{2}{\metre} show that the parallel-wire monopedal robot is capable of achieving the force and velocity required for a jump of \SI{1.6}{\metre}.
  This value of \SI{1.6}{\metre} is 1.45 times higher than the maximum CoG jump height \SI{1.1}{\metre} \cite{haldane2017repetitive, kau2019stanford} of existing leg robots capable of controlled continuous jumping. 
  Furthermore, in order to match the conditions of comparison with the robots that achieved the maximum CoG jump height \SI{1.1}{\metre}\cite{haldane2017repetitive, kau2019stanford}, we consider installing a battery in RAMIEL to power it internally, The total weight is expected to increase from \SI{10}{\kilogram} to \SI{12.5}{\kilogram} \footnote{Three LiPo batteries for motor (6s, \SI{0.5}{\kilogram}), one LiPo battery for circuit power (3s, \SI{0.25}{\kilogram}), and power supply circuit (\SI{0.75}{\kilogram}) for total \SI{2.5}{\kilogram}}.
  If the weight of RAMIEL is increased by a factor of 1.25, the CoG jump height is predicted to be \SI{1.44}{\metre} based on the law of conservation of energy, which is also higher than the maximum CoG jump height \SI{1.1}{\metre} of existing leg robots capable of continuous jumping.
  It is possible that the internal power supply will increase the drop of the power supply voltage at the time of jumping, and the jumping height may be lower than this prediction. 
  However, as described in the \subsecref{subsec:wiremodule}, the theoretical CoG jump height is \SI{5.8}{\metre} when the maximum current of \SI{50}{\ampere} is applied to the motor at the power supply voltage\SI{70}{\volt}. 
  Therefore, it can be inferred that CoG jump height \SI{1.44}{\metre} is feasible even with a slight drop in the power supply voltage.

  In the experiment with the commanded CoG jump height of \SI{1}{\metre}, the robot successfully landed from the height of \SI{0.7}{\metre} without any damage, which confirms the high impact resistance of the parallel wire-driven leg structure due to the lightness of the leg and the back-drivability of the wire.

  On the other hand, the actual jumping height was lower than the commanded CoG jumping height by \SI{0.3}{\metre}-\SI{0.4}{\metre}.
  This is thought to be caused by the fact that the actual tension is lower than the commanded tension due to the friction between the wire and the transit point in the wire module.
  As a countermeasure to the wire friction, the bearing with which the wire comes into contact can be replaced with one with a larger outer diameter.
  In the experiment with a commanded CoG jump height of \SI{2}{\metre}, the wire was cut in midair. 
  This is thought to be caused by repeated contact between the wire and the edge of some part of the leg.
  The countermeasures against wire breakage include fillet the parts that the wire may contact and protect the wire with a bellows-like cover.
}%
{%
  はじめに指令重心跳躍高さ\SI{1}{\metre}の跳躍実験を行う. 
  跳躍時のRAMIELを\figref{figure:20220114onehop1mpicture}に示す.
  RAMIELの本体が水平に接地した状態から跳躍を行い, 跳躍後に開始時の姿勢に復帰していることが分かる.
  RAMIELの重心の垂直方向の変位を\figref{figure:20220114onehop1mtz}に示す。
  離陸時のRAMIELの重心の高さが\SI{0.9}{\metre}であり, 最大高さが\SI{1.63}{\metre}であることからRAMIELの重心が垂直方向に最大\SI{0.73}{\m}変位していることがわかる。
  なお指令重心跳躍高さ\SI{1}{\metre}の跳躍実験における重心の垂直方向の変位はモーションキャプチャを用いて測定した. 

  つぎに指令重心跳躍高さ\SI{2}{\metre}の跳躍実験を行う. 
  跳躍時のRAMIELを\figref{figure:20220201onehop2mpicture}に示す.
  跳躍に成功している一方で, 着地に失敗していることが分かる. 
  また上側のワイヤのうち一本が空中で切断し, 他のワイヤも空中で緩んでいることが分かる. 
  RAMIELの重心の垂直方向の変位を\figref{figure:20220201onehop2mtz}に示す。
  ただし, この指令重心跳躍高さ\SI{2}{\metre}の跳躍実験ではRAMIELがモーションキャプチャの画角を超えてしまうため, 重心の垂直方向の変位をRAMIEL本体に取り付けられたIntel \textregistered  RealSense T265を用いて計測する。
  RealSenseを用いて重心位置を計測しているため, モーションキャプチャを使用した際と比べて重心高さのオフセットが存在する.
  重心跳躍高さは足先が地面から離れた瞬間の重心の高さと跳躍中の最高点での重心の高さの差であるため, 重心高さのオフセットによらず, RealSenseとモーションキャプチャのどちらを用いても同一の重心跳躍高さが計測される. %RealSenseで測定したかモーションキャプチャで測定したかは結果に影響しない
  離陸時のRAMIELの重心の高さが\SI{0.57}{\metre}であり, 最大高さが\SI{2.16}{\metre}であることからRAMIELの重心が垂直方向に最大\SI{1.59}{\m}変位していることがわかる。
  
  これらの実験結果について考察を行う.
  指令重心跳躍高さ\SI{2}{\metre}の実験結果からパラレルワイヤ一本脚ロボットが重心跳躍高さ\SI{1.6}{\metre}の跳躍に必要な力と速度を実現可能であることが分かった.
  重心跳躍高さ\SI{1.6}{\metre}という値は, 既存の連続跳躍可能な脚ロボットの最大重心跳躍高さ\SI{1.1}{\metre} \cite{haldane2017repetitive, kau2019stanford}の1.45倍となっており, 優位に高いことが確認できる.
  さらに最大重心跳躍高さ\SI{1.1}{\metre}を達成したロボットたち\cite{haldane2017repetitive, kau2019stanford}と比較条件を一致させるため, RAMIELにバッテリーを搭載し内部電源化することを考えると, 総重量が\SI{10}{\kilogram}から\SI{2.5}{\kilogram}ほど増加し\footnote{モータ駆動用LiPoバッテリー(6s, \SI{0.5}{\kilogram})3つ, 回路電源用LiPoバッテリー(3s, \SI{0.25}{\kilogram})1つ, 電源回路(\SI{0.75}{\kilogram})で合計\SI{2.5}{\kilogram}という試算}1.25倍の\SI{12.5}{\kilogram}程度になると予想される. 
  RAMIELの重量が1.25倍になると, エネルギー保存則から重心跳躍高さは0.8倍の\SI{1.44}{\metre}となることが予測されるが, この値も既存の連続跳躍可能な脚ロボットの最大重心跳躍高さ\SI{1.1}{\metre}を上回っている.
  %%\subsecref{wiremodule}で述べたように電源電圧\SI{70}{\volt}で\SI{50}{\ampere}の最大電流をモータに印加した状態での理論上の重心跳躍高さは\SI{5.8}{\metre}であり, 内部電源化したことにより電源電圧がより大きく降下するようになったとしても.
  内部電源化することで跳躍時の電源電圧の降下幅が大きくなり, 跳躍高さがこの予想より低くなる可能性も考えられる. 
  しかし\subsecref{subsec:wiremodule}で述べたように電源電圧\SI{70}{\volt}で\SI{50}{\ampere}の最大電流をモータに印加した状態での理論上の重心跳躍高さが\SI{5.8}{\metre}であるため, 電源電圧が多少降下しても重心跳躍高さ\SI{1.44}{\metre}の跳躍は十分実現可能であると推測できる.

  また指令重心跳躍高さ\SI{1}{\metre}の実験では高さ\SI{0.7}{\metre}からの着地に成功しさらに破損もなかったことから, 脚の軽量性やワイヤのバックドライバビリティに起因するパラレルワイヤ脚構造の耐衝撃性の高さが確認できた.
  一方で指令重心跳躍高さに対し実際に跳躍した高さが\SI{0.3}{\metre}-\SI{0.4}{\metre}ほど低くなっている.
  これはワイヤがワイヤモジュール内部の経由点などと摩擦を生じ, 実際の張力が指令張力より低くなっていることが原因であると考えられる.
  ワイヤに生じる摩擦への対策としては, ワイヤが接触するベアリングを外径がより大きいものに交換することがなどが挙げられる.
  そして指令重心跳躍高さ\SI{2}{\metre}の実験では空中でワイヤが切断された. 
  これは脚の何らかのパーツのエッジとワイヤが接触を繰り返し摩耗したことで発生したと考えられる.
  ワイヤの切断への対策としては, ワイヤが接触しうるパーツにフィレットをかけること, ワイヤに蛇腹状のカバーをつけて保護することなどが挙げられる.
}%

\subsection{Continuous Jump}
\begin{figure}[t]
  \begin{center}
   \includegraphics[width=1.0\columnwidth]{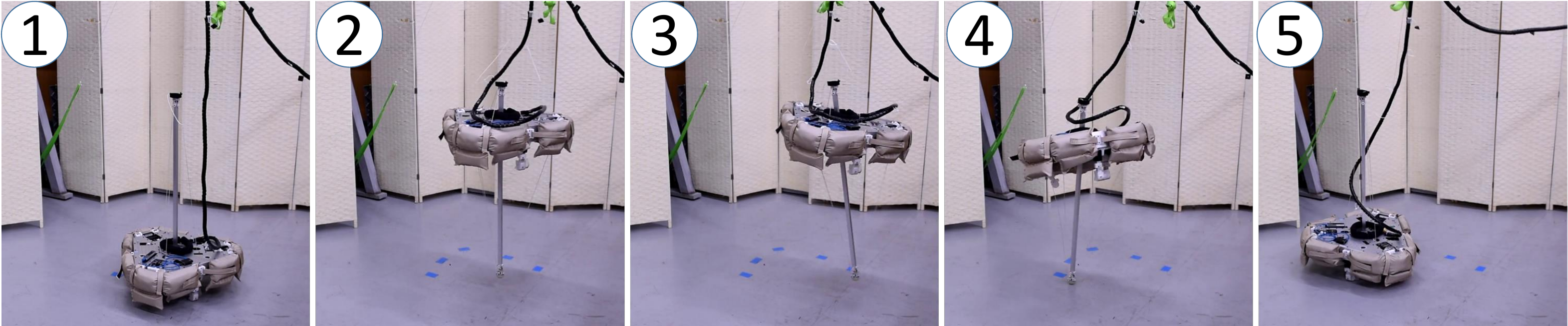}
   \caption{Snapshots of 7 continuous jumping motion. RAMIEL can jump and land from seated position.}
   \label{figure:20220113eighthop11picture}
  \end{center}
\end{figure}

\begin{figure}[tbh]
  \begin{center}
   \includegraphics[width=1.0\columnwidth]{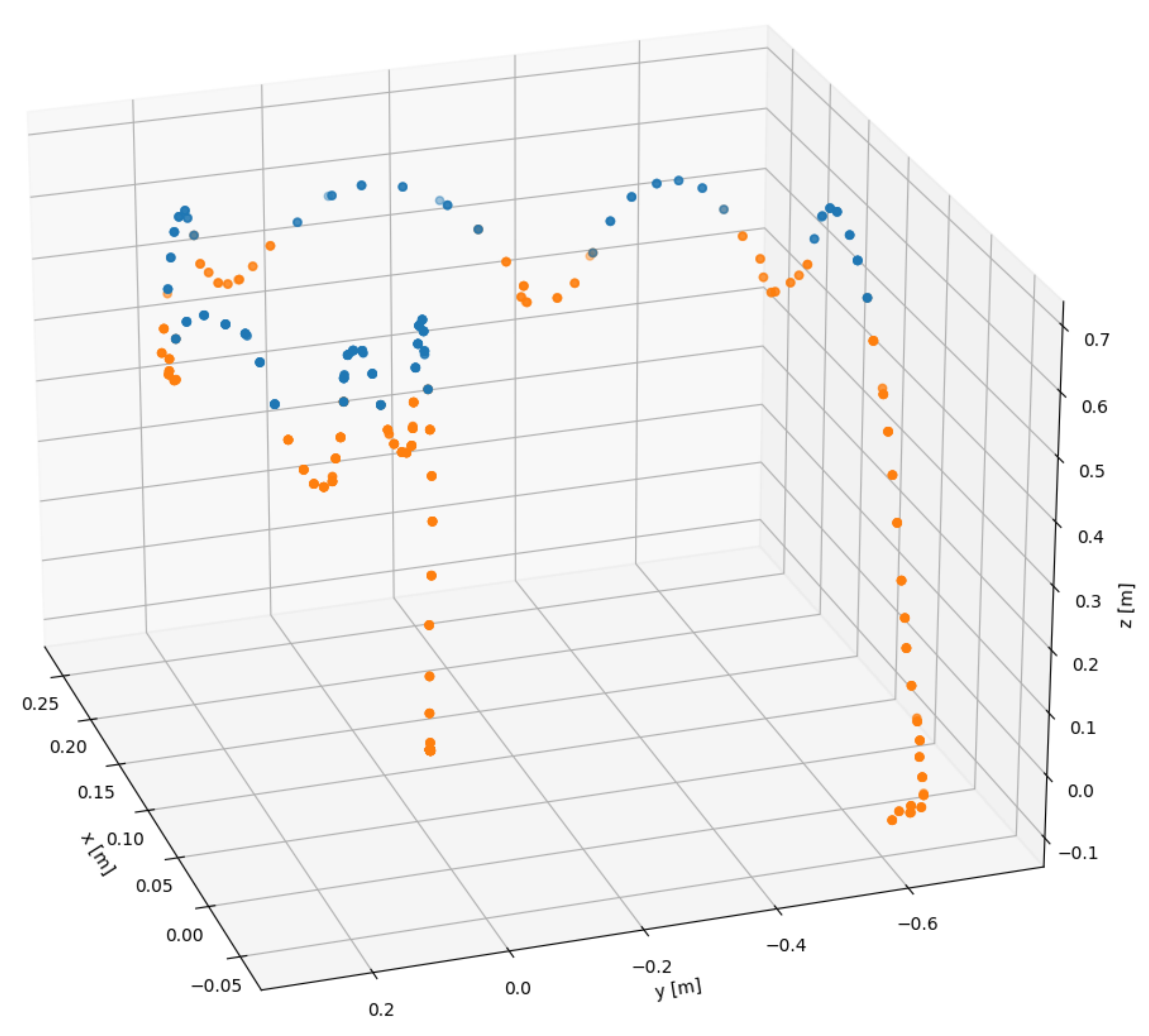}
   \caption{Position of RAMIEL's center of gravity (CoG) during 7 continuous jumping motion. Stance phases are shown in orange. Flight phases are shown in blue.}
   \label{figure:20220113eighthop11xyz}
  \end{center}
\end{figure}

\switchlanguage%
{%
  In this section, we perform continuous jumping experiments using RAMIEL. 
  In this experiment, the CoG of RAMIEL is measured by using RealSense T265 installed in RAMIEL itself.
  
  The results of 16 trials for continuous jumping experiments with the same control parameters and environment are summarized in \tabref{table:continuousjump}.
  In 5 of these trials, the robot succeeded in jumping more than 5 consecutive times (The number of times the leg left the ground is considered to be the number of continuous jumps).
  The results of the 14th jump experiment are shown in \figref{figure:20220113eighthop11picture}.
  The position of the CoG of RAMIEL during the experiment is shown in \figref{figure:20220113eighthop11xyz}.
  From these figures, it can be confirmed that RAMIEL made seven continuous jumps from the seated state and returned to the seated state after landing.
  These experimental results show that the parallel wire-driven leg structure has high controllability of the leg required for continuous jumping. 

  On the other hand, in nine out of fifteen experiments, the number of continuous jumps ended at two.
  In these experiments, the leg was tilted so much in the first stance that they gained horizontal speed, and in the second stance, the leg was tilted so much that they exceeded the range of motion of the joints.
  One of the causes of such falls is that the power cable contacts RAMIEL and generates a disturbance force.
  It is expected that the stability of the robot can be improved by installing a battery in the robot to eliminate the disturbance of the power cable and by adding an ankle to increase the DoF of the control.
}%
{%
  本節ではRAMIELを用いて連続跳躍実験を行う. 
  なお本実験ではRAMIELの重心位置の測定をRAMIEL本体に設置されたRealSense T265を用いて行う。
  
  同じ制御パラメータ、環境で合計17回ほど連続跳躍実験を行った結果を\tabref{table:continuousjump}にまとめる。
  このうち5連続以上の跳躍に成功したのは17回中5回であった。
  ただし脚が地面から離れた回数を連続跳躍の回数とする。

  14回目の連続跳躍実験の結果を\figref{figure:20220113eighthop11picture}に示す。
  またその実験時のRAMIELの重心位置を\figref{figure:20220113eighthop11xyz}に示す。
  これらの図から着座状態から7回連続で跳躍を行い着地後に着座状態に復帰していることが確認できる。
  この実験結果からパラレルワイヤ脚構造が連続跳躍に求められる脚の高い制御性を有していることが分かった. 

  一方で15回中9回の実験において連続跳躍回数が2回で終了している.
  これらの回では一回目の立脚器に脚が大きく傾いて水平方向の速度がつき、2回目の遊脚期に機体が大きく傾き関節の可動範囲を超えたことでバランスを崩し転倒するケースが多かった.
  このような転倒の原因としては, 電源ケーブルがRAMIELと接触し外乱となる力を発生させていることが挙げられる.
  ロボットにバッテリーを搭載し外乱となる電源ケーブルを排除することや, 足首を付与し制御の自由度を増加させることなどにより安定性が向上し連続跳躍回数が増加することが込まれる.

  %連続跳躍実験データまとめシート: https://docs.google.com/spreadsheets/d/1Yvz0Jla5SZs7GKKhakS3st_q8TRIrVFcHDq10yykH0I/edit?usp=share_link
}%

\begin{table*}[t]
  \begin{center}
  \caption{Continuous jumping experimental results. The number of times the leg leave the ground is the number of continuous jumps.}
    \begin{tabular}{ccccccccccccccccc} \toprule
      Trial & 1 & 2 & 3 & 4 & 5 & 6 & 7 & 8 & 9 & 10 & 11 & 12 & 13 & 14 & 15 & 16 \\ \midrule
      Number of continuous jumps & 2 & 2 & 2 & 3 & 8 & 5 & 2 & 2 & 2 & 2 & 6 & 2 & 0 & 7 & 2 & 6 \\ \bottomrule
    \end{tabular}
    \label{table:continuousjump}
  \end{center}
\end{table*}

\section{Conclusion} \label{sec:conclusion}
\switchlanguage%
{%
  In this study, we proposed a parallel wire-driven leg structure in order to clarify the leg structure which can realize continuous jumping and high jumping. 
  The parallel wire-driven leg structure has DoFs for posture control, which enable continuous jumping, and the linear motion joint is more advantageous for high jumping than the rotary joint. 
  We developed a parallel-wire monopedal robot, RAMIEL, which has both the important conditions for high jumping, i.e., high velocity at take-off, and the important condition for continuous jumping, i.e., light weight leg. 
  Furthermore, we have achieved a CoG jump height of \SI{1.6}{\metre} and a maximum of seven continuous jumps with RAMIEL.
  From these experiments, we concluded that the parallel wire-driven leg structure can realize continuous jumps and high jumps. 
  
  In the future, we will add an encoder to each joint to improve the accuracy of joint angle estimation and add an ankle to control the posture of the leg during stance. 
}%
{%
  本研究では連続跳躍と高い跳躍を実現する脚構造を明らかにすることを目的とし, パラレルワイヤ脚構造を提案した. 
  パラレルワイヤ脚構造には姿勢制御用の自由度があり連続跳躍が可能であること, 直動関節を採用したことで回転関節に比べて高い跳躍に有利であることを述べた. 
  そして離陸時の速度が大きいという高い跳躍を行う上で重要な条件と, 脚が軽量であるという連続跳躍を行う上で重要な条件を両立するパラレルワイヤ一本脚ロボットRAMIELを開発した. 
  さらにRAMIELを用いて重心跳躍高さ\SI{1.6}{\metre}の跳躍と, 最大7回の連続跳躍を実現した.
  これらの実験からパラレルワイヤ脚構造が連続跳躍と高い跳躍を実現する脚構造であるという結論を得た. 
  
  今後は各関節にエンコーダを追加し関節角度推定の精度を向上させることと足首を付与し立脚時に脚の姿勢を制御可能にすることを行う. 
}%

{
  %\footnotesize
  %\small
  %\bibliographystyle{junsrt}
  \bibliographystyle{IEEEtran}
  \bibliography{main}
}

\end{document}